\documentclass[twoside]{article}

\usepackage[accepted]{aistats2026}

\usepackage[ruled,vlined,linesnumbered]{algorithm2e}
\usepackage{amsfonts}
\usepackage{amsmath}
\usepackage{amsthm}
\usepackage{array}
\usepackage{bbm}
\usepackage{booktabs}
\usepackage{caption}
\usepackage{dsfont}
\usepackage{enumitem}
\usepackage[T1]{fontenc}
\usepackage[pdftex]{graphicx}
\usepackage[pagebackref]{hyperref}
\usepackage[utf8]{inputenc}
\usepackage{makecell}
\usepackage{mathtools}
\usepackage{microtype}
\usepackage{multirow}
\usepackage[authoryear,round]{natbib}
\usepackage{nicefrac}
\usepackage{pifont}
\usepackage{ragged2e}
\usepackage{stmaryrd}
\usepackage{subcaption}
\usepackage{tabularx}
\usepackage{tcolorbox}
\usepackage{thm-restate}
\usepackage{tikz}
\usepackage{url}
\usepackage{xcolor}
\usepackage{xspace}
\usepackage{xurl}

\usepackage[capitalise]{cleveref}

\newcommand{\cbr}[1]{\left\{ #1 \right\}}
\newcommand{\nbr}[1]{\lvert #1 \rvert}
\newcommand{\rbr}[1]{\left( #1 \right)}
\newcommand{\sbr}[1]{\left[ #1 \right]}

\newcommand{\ind}[1]{\mathds{1} \cbr{#1}}
\newcommand{\pr}[1]{\mathbf{P} \cbr{#1}}
\newcommand{\ev}[1]{\mathbf{E} \sbr{#1}}

\SetKwComment{Comment}{/* }{ */}
\newcolumntype{C}[1]{>{\centering\arraybackslash}p{#1}}
\newcolumntype{L}[1]{>{\raggedright\arraybackslash}p{#1}}

\usetikzlibrary{chains}

\hypersetup{
    colorlinks,
    linkcolor={red!50!black},
    citecolor={blue!50!black},
    urlcolor={blue!80!black}
}

\begin{document}

\newcommand{\papertitle}{E-Scores for (In)Correctness Assessment of Generative Model Outputs}
\runningauthor{Guneet S. Dhillon, Javier Gonz\'{a}lez, Teodora Pandeva, Alicia Curth}

\renewcommand*{\thefootnote}{\fnsymbol{footnote}}

\twocolumn[
    \aistatstitle{\papertitle}
    \aistatsauthor{Guneet S. Dhillon\footnotemark[1] \And Javier Gonz\'{a}lez}
    \aistatsaddress{University of Oxford \\ \texttt{guneet.dhillon@stats.ox.ac.uk} \And Microsoft Research \\ \texttt{jav.gonzalezh@gmail.com}}
    \aistatsauthor{Teodora Pandeva \And Alicia Curth}
    \aistatsaddress{Microsoft Research \\ \texttt{tpandeva@microsoft.com} \And Microsoft Research \\ \texttt{aliciacurth@microsoft.com}}
]

\footnotetext[1]{Work done while at Microsoft Research.}
\renewcommand*{\thefootnote}{\arabic{footnote}}
\setcounter{footnote}{0}

\begin{abstract}
While generative models, especially large language models (LLMs), are ubiquitous in today’s world, principled mechanisms to assess their (in)correctness are limited. Using the conformal prediction framework, previous works construct sets of LLM responses where the probability of including an incorrect response, or error, is capped at a user-defined tolerance level. However, since these methods are based on p-values, they are susceptible to p-hacking, i.e., choosing the tolerance level post-hoc can invalidate the guarantees. We therefore leverage e-values to complement generative model outputs with e-scores as measures of incorrectness. In addition to achieving the guarantees as before, e-scores further provide users with the flexibility of choosing data-dependent tolerance levels while upper bounding size distortion, a post-hoc notion of error. We experimentally demonstrate their efficacy in assessing LLM outputs under different forms of correctness: mathematical factuality and property constraints satisfaction.
\end{abstract}

\section{INTRODUCTION}
\label{sec:intro}

\begin{figure*}[!t]
    \begin{center}
        \begin{small}
            \begin{tikzpicture}[start chain=going below, node distance=0cm]
                \node [on chain, text width=14cm] {
                    \centering
                    \textbf{PROMPT}
                    \begin{tcolorbox}[left=0.1cm, right=0.1cm, top=0.1cm, bottom=0.1cm, colback=black!10!white]
                        \centering
                        Seth is twice as old as Brooke. In 2 years, the sum of their ages will be 28. How old is Seth?
                    \end{tcolorbox}
                };
                \node [on chain, text width=16.9cm] {
                    \centering
                    \textbf{RESPONSE}
                    \begin{tcolorbox}[left=0.1cm, right=0.1cm, top=0.1cm, bottom=0.1cm, colback=red!10!white]
                        \begin{tcolorbox}[left=0.1cm, right=0.1cm, top=0.1cm, bottom=0.1cm, colback=red!10!white]
                            \begin{tcolorbox}[left=0.1cm, right=0.1cm, top=0.1cm, bottom=0.1cm, colback=red!10!white]
                                \begin{tcolorbox}[left=0.1cm, right=0.1cm, top=0.1cm, bottom=0.1cm, colback=green!10!white]
                                    \begin{tcolorbox}[left=0.1cm, right=0.1cm, top=0.1cm, bottom=0.1cm, colback=green!10!white]
                                        \centering
                                        Let's denote Seth's current age as $S$ and Brooke's current age as $B$. From the problem, we know that Seth is twice as old as Brooke, so we can write this relationship as an equation: $S = 2B$.
                                        \tcblower
                                        \ding{51} \hfill \textbf{e-score: 0.01}
                                    \end{tcolorbox}
                                    \centering
                                    In two years, Seth will be $S + 2$ years old and Brooke will be $B + 2$ years old. The sum of their ages in two years will be 28, so we can write another equation: $(S + 2) + (B + 2) = 28$.
                                    \tcblower
                                    \ding{51} \hfill \textbf{e-score: 0.01}
                                \end{tcolorbox}
                                \centering
                                Substituting the first equation into the second equation, we get: $2B + \pmb{4} + B + 2 = 28 \iff 3B + 6 = 28 \iff 3B = 22 \iff B = \nicefrac{22}{3}$.
                                \tcblower
                                \ding{55} \hfill \textbf{e-score: 4.95}
                            \end{tcolorbox}
                            \centering
                            Since $B$ represents Brooke's age, we need to find Seth's age. From the first equation, we substitute $B$ with $\nicefrac{22}{3}$: $S = 2 \times \nicefrac{22}{3} \iff S = \nicefrac{44}{3}$.
                            \tcblower
                            \ding{55} \hfill \textbf{e-score: 6.01}
                        \end{tcolorbox}
                        \centering
                        So, Seth is $\nicefrac{44}{3}$ or approximately $14.67$ years old. The answer is: $14.67$.
                        \tcblower
                        \ding{55} \hfill \textbf{e-score: 6.28}
                    \end{tcolorbox}
                };
            \end{tikzpicture}
        \end{small}
    \end{center}
    \caption{
        \textbf{E-scores example for mathematical factuality.}
        This is an example prompt and response from the ProcessBench benchmark \citep{zheng2025processbench} (cf. \cref{subsec:experiments-processbench}).
        The LLM's response is made up of 5 sub-responses, each a step in the mathematical reasoning (starting from the inner and ending on the outer block).
        The checks/crosses on the bottom left and the green/red colour of each block represent whether the response is correct/incorrect up till that point (which is not known at test-time).
        On manual inspection, we bold part of the third sub-response causing the incorrectness, cascading to subsequent sub-responses.
        The e-scores on the bottom right of each block are our proposed measures of incorrectness: low for correct and high for incorrect responses.
    }
    \label{fig:example}
\end{figure*}
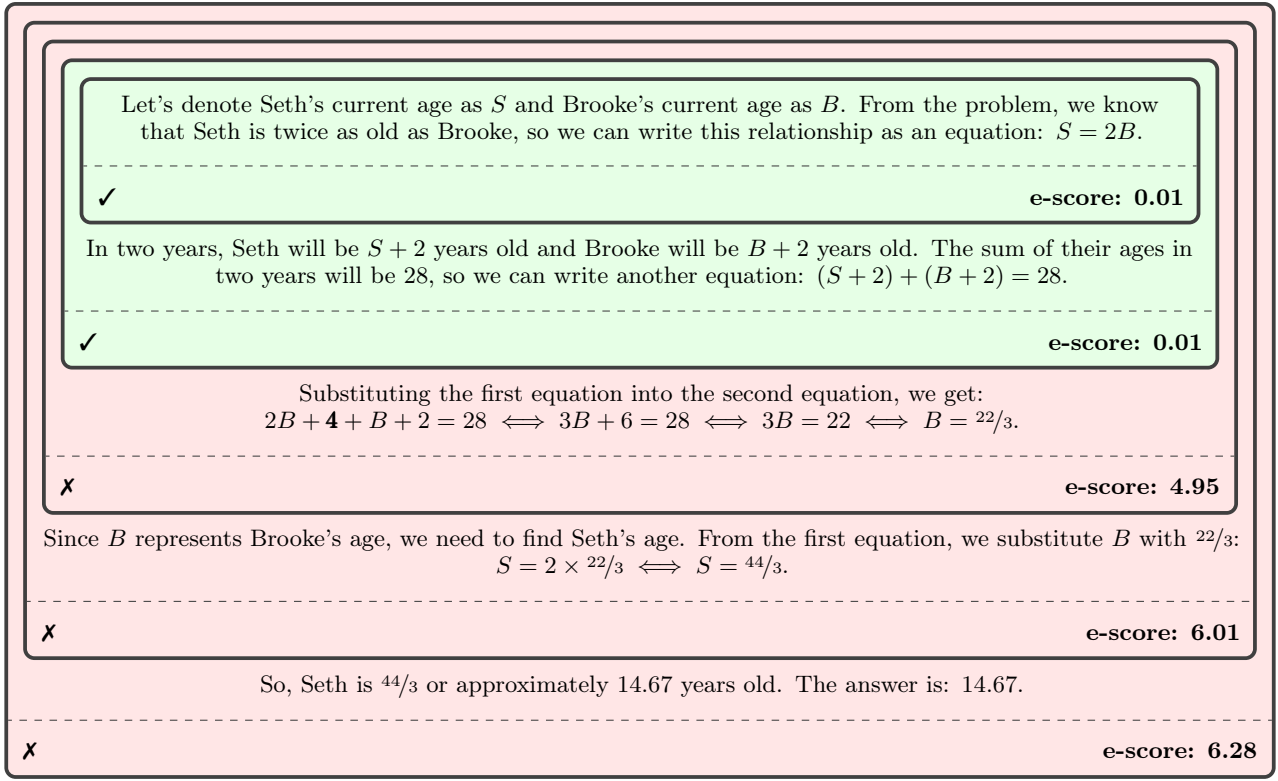

Generative models, large language models (LLMs) in particular, have gained widespread popularity, with millions of users around the world \citep{openai2024gpt4,openai2024gpt4o,openai2024openaio1,gemini2025gemini2.5}.
However, they are susceptible to generating incorrect outputs, or \emph{hallucinations} \citep{huang2025survey}, requiring caution in their use.
\cref{fig:example} illustrates an example where an LLM's response contains incorrect steps or sub-responses.
Since such correctness labels are unknown at test time, we need a mechanism to assess the (in)correctness of the generated responses.

A recent line of work aims to provide statistical guarantees for such LLM responses \citep{mohri2024language,cherian2024large,rubintoles2025conformal}.
These methods rely on \emph{p-value} based conformal prediction \citep{shafer2008tutorial,vovk2022algorithmic} to filter the response set such that the probability of including an incorrect response, or error, is capped at a user-defined tolerance level $\alpha$.
Implicitly, this is done by computing a p-value based score for each response, or \emph{p-score}; then, the filtered response set is obtained simply by thresholding the p-scores $\leq \alpha$.
Importantly, $\alpha$ must be chosen independently of the data.
This begs the question: \emph{what if we want a data-dependent $\alpha$?}

Recall the example in \cref{fig:example}, and imagine that the scores there are the p-scores in question.
If the user pre-sets $\alpha = 0.1$, the responses are filtered for p-scores $\leq 0.1$, resulting in the first two sub-responses.
However, the user would have obtained the same first two sub-responses if they had pre-set $\alpha = 0.01$ instead.
Since a tolerance level of $0.01$ conveys a much higher assurance in the responses compared to $0.1$, the user would want to update $\alpha = 0.1 \rightarrow 0.01$.
This necessitates a data-dependent $\alpha$, also called a \emph{post-hoc $\alpha$}.
Unfortunately, even though such post-hoc $\alpha$'s are commonly used in practice, the guarantees for p-score based methods are invalidated.
This is due to the susceptibility of p-values, and hence of p-scores, to \emph{p-hacking} \citep{carney2016position}.

We propose \emph{e-scores} as \emph{measures of incorrectness}: they are low for correct and high for incorrect responses (depicted in \cref{fig:example}).
These scores, based on \emph{e-values}, provide statistical guarantees on a post-hoc notion of error called \emph{size distortion} \citep{koning2024posthoc}: the distortion between observing an error and the user's post-hoc tolerance level.
The non-post-hoc error guarantee mentioned earlier arises as a special case of this generalization.
Furthermore, we show that our statistical guarantees remain valid for any generative model and for a super-set of the response sets considered by \citet{mohri2024language,cherian2024large,rubintoles2025conformal}.
Altogether, this provides avenues for more diverse applications and use-cases.
We experimentally demonstrate the efficacy of our e-scores in the assessment of LLM responses under two settings: (i) mathematical factuality, for sound mathematical reasoning, and (ii) property constraints satisfaction, to ensure responses satisfy certain desirable properties.

We summarize the contributions of our work as follows.
\begin{itemize}
    \item
    We study the problem of achieving statistical guarantees for a post-hoc notion of error, namely size distortion, for generative model outputs.
    This generalizes the non-post-hoc guarantees studied before for LLM responses \citep{mohri2024language,cherian2024large,rubintoles2025conformal}.
    
    \item
    We propose e-scores, based on e-values, as measures of incorrectness.
    Our theoretical results show that e-scores achieve the post-hoc statistical guarantees mentioned above.
    In doing so, e-scores provide users the flexibility of choosing data-dependent tolerance levels.
    Furthermore, we corroborate our theory with experimental results.
    
    \item
    We show that our guarantees extend to any generative model and to a super-set of the response sets considered by \citet{mohri2024language,cherian2024large,rubintoles2025conformal}.
\end{itemize}

We begin by formulating our problem in \cref{sec:setup}, and discuss related work in \cref{sec:related}.
We define our proposed e-scores in \cref{sec:e_scores}.
\cref{sec:experiments,sec:theory} include experimental results and theoretical analyses, respectively.
We finish with concluding remarks in \cref{sec:conclusions}.

\section{PROBLEM FORMULATION}
\label{sec:setup}

We are interested in providing guarantees pertaining to the (in)correctness of generative model outputs.
We describe: (i) the generative model outputs we consider, (ii) their (in)correctness, (iii) the desired post-hoc guarantees, and (iv) practical examples of post-hoc use.
While we use LLMs to provide concrete examples, our setup could be instantiated with any generative model.

\subsection{Generative Model Outputs}

We define the prompt space $\mathcal{X}$, a sub-space of language.
Given a prompt $x \in \mathcal{X}$, an LLM $\pi$ generates a response,
\begin{equation*}
    g_{\pi} \rbr{x} = \mathbf{y} = \rbr{y_{1} , y_{2} , \ldots} \sim \pi \rbr{\cdot | x} ,
\end{equation*}
an ordered set of sub-responses that collectively answer the prompt.
These sub-responses could be sentences, steps in chain-of-thought reasoning \citep{wei2022chainofthought}, etc., akin to the example in \cref{fig:example}.
While natural for auto-regressive models, we do not assume any particular dependency structure.
Note that singular responses of length $\nbr{\mathbf{y}} = 1$ are a special case.
We define the sub-response space $\mathcal{Y}$, also a sub-space of language.
Then, each sub-response $y_{i} \in \mathcal{Y}$ and the response $\mathbf{y} \in \cup_{i \geq 1} \mathcal{Y}^{i}$.

We could evaluate $\mathbf{y}$ alone, but we consider a larger set of responses.
Notice that a \emph{single} generated response can form \emph{multiple} responses: the partial responses $\mathbf{y}_{\leq i} = \rbr{y_{1} , \ldots , y_{i}}$, for $i = 1 , \ldots , \nbr{\mathbf{y}}$.
In doing so, the user could find partial but correct responses, similar to the example in \cref{fig:example}.
We define such a response set,
\begin{equation}
    \label{eq:simple_response_set}
    \mathds{Y} \rbr{g_{\pi} \rbr{x}} = \cbr{\mathbf{y}_{\leq i} : \mathbf{y} = g_{\pi} \rbr{x} , i = 1 , \ldots , \nbr{\mathbf{y}}} .
\end{equation}
Note that this includes the full generated response itself $g_{\pi} \rbr{x} \in \mathds{Y} \rbr{g_{\pi} \rbr{x}}$.
We will generalize this response set to a larger one in \cref{subsec:theory-setup}.
However, we continue to use the definition in \cref{eq:simple_response_set} as it is suited for the experimental benchmarks we consider (cf. \cref{sec:experiments}).

\subsection{Oracle (In)Correctness}

There are different notions of \emph{correctness} that are of interest to us.
Factuality is a pertinent one, to verify whether responses are based on facts \citep{mohri2024language,cherian2024large,rubintoles2025conformal}.
Another is property constraints satisfaction, to ensure responses have certain desirable properties \citep{dhillon2025l3ms}.
To adapt to such use-cases, we define (in)correctness \emph{with respect to an oracle $o$},
\begin{equation*}
    o \rbr{x , \mathbf{y}} = \begin{cases}
        1 , & \text{$\mathbf{y}$ is correct as a response to $x$} \\
        0 , & \text{$\mathbf{y}$ is incorrect as a response to $x$}
    \end{cases} ,
\end{equation*}
determining the correctness of $\mathbf{y}$ as a response to the prompt $x$.
Further, we define the labeled response set,
\begin{equation*}
    \mathds{O} \rbr{x , g_{\pi} \rbr{x}} = \cbr{\rbr{\mathbf{y} , o \rbr{x , \mathbf{y}}} : \mathbf{y} \in \mathds{Y} \rbr{g_{\pi} \rbr{x}}} .
\end{equation*}

\subsection{Desideratum for Statistical Guarantees}

Given an LLM $\pi$ and a prompt $x \in \mathcal{X}$, we construct its (unlabeled) response set $\mathds{Y} \rbr{g_{\pi} \rbr{x}}$.
Our goal then is to assess the (in)correctness of each response in this set, i.e., we want to reason about the unknown oracle labels $o \rbr{x , \mathbf{y}}$ for every response $\mathbf{y} \in \mathds{Y} \rbr{g_{\pi} \rbr{x}}$.
We do so by complementing each response with a non-negative score $s \rbr{x , \mathbf{y}} \in \mathds{R}_{\geq 0}$ as a \emph{measure of incorrectness}: low for correct responses and high for incorrect responses.
Consequently, we will provide the scored response set,
\begin{equation*}
    \mathds{S} \rbr{x , g_{\pi} \rbr{x}} = \cbr{\rbr{\mathbf{y} , s \rbr{x , \mathbf{y}}} : \mathbf{y} \in \mathds{Y} \rbr{g_{\pi} \rbr{x}}} ,
\end{equation*}
to facilitate the user in deciding which responses to include (and use) or not.
In particular, they could decide to filter the scored response set at some $\alpha \in \mathds{R}_{\geq 0}$,
\begin{equation*}
    \mathds{S}_{\alpha} \rbr{x , g_{\pi} \rbr{x}} = \cbr{\rbr{\mathbf{y} , v} \in \mathds{S} \rbr{x , g_{\pi} \rbr{x}} : v \leq \alpha} .
\end{equation*}

Since we want to avoid incorrect responses, we treat the inclusion of any incorrect response in the filtered set $\mathds{S}_{\alpha} \rbr{x , g_{\pi} \rbr{x}}$ as an \emph{error at $\alpha$}.
Then, a possible desideratum for our measures of incorrectness is to upper bound the probability of error at $\alpha$ by $\alpha$ itself,
\begin{equation}
    \label{eq:size}
    \pr{\text{error at $\alpha$}} \! = \! \pr{\splitfrac{\exists \rbr{\mathbf{Y} , \cdot} \in \mathds{S}_{\alpha} \rbr{X , g_{\pi} \rbr{X}}}{\text{ s.t. } o \rbr{X , \mathbf{Y}} = 0}} \! \leq \! \alpha .
\end{equation}
In doing so, $\alpha$ represents the user's tolerance level.
This is considered by \citet{mohri2024language,cherian2024large,rubintoles2025conformal}, who use p-value based conformal prediction \citep{shafer2008tutorial,vovk2022algorithmic} to achieve this.
Note that the above requirement assumes that $\alpha$ is determined independently of the data (the prompt, the responses, and the scores).
However, in practice, users would want to use a data-dependent tolerance level $\alpha$, as highlighted by the scenario in \cref{sec:intro}.
On realizing that they would obtain the same filtered set if they pre-set $\alpha = 0.01$ instead of $0.1$ in the example in \cref{fig:example}, the user wants to update $\alpha = 0.1 \rightarrow 0.01$, conveying higher assurance in the responses with a smaller tolerance level.
This necessitates a data-dependent $\alpha$, or a \emph{post-hoc $\alpha$}.

Specifically, we want to enable the user to choose $\alpha$ after observing the prompt $x$ and the scored response set $\mathds{S} \rbr{x , g_{\pi} \rbr{x}}$.
Since $\alpha$ is now a random variable, \cref{eq:size} cannot be applied directly.
We therefore generalize our desideratum instead to a post-hoc notion of error at $\alpha$,
\begin{equation}
    \label{eq:size_distortion}
    \begin{gathered}
        \ev{\frac{\ind{\text{error at $\alpha \rbr{X , \mathds{S} \rbr{X , g_{\pi} \rbr{X}}}$}}}{\alpha \rbr{X , \mathds{S} \rbr{X , g_{\pi} \rbr{X}}}}} = \\
        \ev{\frac{\ind{\splitfrac{\! \exists \! \rbr{\mathbf{Y} , \cdot} \! \in \! \mathds{S}_{\alpha \rbr{X , \mathds{S} \rbr{X , g_{\pi} \rbr{X}}}} \rbr{X , g_{\pi} \rbr{X}}}{\! \text{s.t. } \! o \rbr{X , \mathbf{Y}} \! = \! 0}}}{\alpha \rbr{X , \mathds{S} \rbr{X , g_{\pi} \rbr{X}}}}} \! \leq \! 1 .
    \end{gathered}
\end{equation}
The ratio of observing an error at $\alpha$ and $\alpha$ itself is expected to be at most 1.
The ratio captures the distortion between observing an error and the user's tolerance level; furthermore, the bound ensures that the expected distortion is controlled.
\citet{koning2024posthoc} uses such an expected distortion for hypothesis testing, calling it \emph{size distortion} (with size referring to an error).
This generalizes \cref{eq:size}, recovered when $\alpha$ is a fixed pre-set value.
Thus, \cref{eq:size_distortion} will be our new desideratum.

\paragraph{Role of the Calibration Data}
To aid us in our desideratum, we are given labeled calibration data that is assumed to be exchangeable with the test data, which is similar to \citet{mohri2024language,cherian2024large,rubintoles2025conformal}.
This includes $n$ calibration prompts $x^{i} \in \mathcal{X}$, for $i = 1 , \ldots , n$, each with their corresponding labeled response set $\mathds{O} \rbr{x^{i} , g_{\pi} \rbr{x^{i}}}$.
Now, given the test prompt $x^{n + 1} \in \mathcal{X}$, we will complement each test response $\mathbf{y}^{n + 1} \in \mathds{Y} \rbr{g_{\pi} \rbr{x^{n + 1}}}$ with a non-negative test score $s \rbr{x^{n + 1} , \mathbf{y}^{n + 1}} \in \mathds{R}_{\geq 0}$ as a measure of incorrectness, which can additionally depend on the given calibration data.
For simplicity and compactness, we will suppress the notation for the dependence of the scores on the calibration data; note that the requirement in \cref{eq:size_distortion} (and in \cref{eq:size}) is now \emph{marginal over both the test and the calibration data}.

\subsection{Use of Post-Hoc \texorpdfstring{$\pmb{\alpha}$}{alpha}'s}
\label{subsec:alpha_strategies}

We provide two concrete examples of post-hoc $\alpha$ strategies.
A user could choose either one of these or any other strategy; in return, we will aim to satisfy \cref{eq:size_distortion}.

\paragraph{Max-Constrained Adaptive $\pmb{\alpha}$}
The user has a fixed pre-set maximum tolerance level $\alpha_{\max} \in \sbr{0 , 1}$.
Since the scores in the filtered set at $\alpha_{\max}$ could be $\leq \alpha_{\max}$, the user updates their tolerance level $\alpha$ to the maximum score in the corresponding filtered set $\mathds{S}_{\alpha_{\max}} \rbr{x , g_{\pi} \rbr{x}}$,
\begin{equation*}
    \alpha \rbr{x , \mathds{S} \rbr{x , g_{\pi} \rbr{x}}} = \max_{\rbr{\cdot , v} \in \mathds{S}_{\alpha_{\max}} \rbr{x , g_{\pi} \rbr{x}}} v .
\end{equation*}

\paragraph{Fractional Inclusion}
Alternatively, the user might want to include $\lambda \in \sbr{0 , 1}$ fraction of all responses in the set $\mathds{S} \rbr{x , g_{\pi} \rbr{x}}$.
Then, the user's tolerance level $\alpha$ is the maximum score in the corresponding filtered set,
\begin{equation*}
    \begin{gathered}
        \alpha \rbr{x , \mathds{S} \rbr{x , g_{\pi} \rbr{x}}} = \max_{\rbr{\cdot , v} \in \mathds{S}_{\alpha} \rbr{x , g_{\pi} \rbr{x}}} v \\
        \text{s.t. } \nbr{\mathds{S}_{\alpha} \rbr{x , g_{\pi} \rbr{x}}} = \lceil \lambda \cdot \nbr{\mathds{S} \rbr{x , g_{\pi} \rbr{x}}} \rceil .
    \end{gathered}
\end{equation*}
Note that by setting $\lambda = 1$, the user could get post-hoc error guarantees for the full generated response itself.

After choosing a post-hoc $\alpha$ strategy, the user gets the filtered $\mathds{S}_{\alpha \rbr{x , \mathds{S} \rbr{x , g_{\pi} \rbr{x}}}} \rbr{x , g_{\pi} \rbr{x}}$ for downstream use.
The user could, for example, treat the longest response in the filtered set as the default response, as done by \citet{mohri2024language,cherian2024large,rubintoles2025conformal}, now with added guarantees.

\section{RELATED WORK}
\label{sec:related}

We begin with two key definitions.
Consider a non-negative random variable $R \in \mathds{R}_{\geq 0}$.
It is a \emph{p-variable} if $\pr{R \leq \alpha} \leq \alpha$, for all $\alpha \in \mathds{R}_{\geq 0}$.
And, it is an \emph{e-variable} if $\ev{R} \leq 1$ (which, with Markov's inequality, gives $\pr{\nicefrac{1}{R} \leq \alpha} \leq \alpha$, for all $\alpha \in \mathds{R}_{\geq 0}$).
Furthermore, its realized value is called a p- and e-value, respectively.

While p-values have been used for hypothesis testing \citep{neyman1933problem,wald1939contributions}, recent developments highlight the benefits of e-values \citep{shafer2019gametheoretic,wasserman2020universal,shafer2021testing,vovk2021evalues,wang2022false,grunwald2024safe,ramdas2025hypothesis}.
Notably, \citet{grunwald2024beyond} emphasizes their use in post-hoc $\alpha$ settings.
Since we are also interested in post-hoc $\alpha$'s, we base our scores on e-values to attain statistical guarantees.

Closest to our work is that of \citet{mohri2024language,cherian2024large,rubintoles2025conformal}.
\citet{mohri2024language,rubintoles2025conformal} adapt ideas from conformal prediction \citep{shafer2008tutorial,vovk2022algorithmic}, typically used to construct prediction sets for supervised learning problems, to filter LLM outputs to construct response sets $\mathds{S}_{\alpha} \rbr{x , g_{\pi} \rbr{x}}$ for a fixed pre-set $\alpha$.
In both these works, the dependence on p-values is implicit through their use of the nested conformal framework \citep{gupta2022nested}.
Additionally, rather than a single fixed $\alpha$, \citet{cherian2024large} consider a functional $\alpha$ that can vary to improve fractional inclusion, but is required to be independent of the scores.
Consequently, these works satisfy \cref{eq:size}, but not its post-hoc generalization in \cref{eq:size_distortion}.
We therefore design our scores to achieve the latter for any generative model and its outputs.
Furthermore, our theoretical results extend to the assessment of response sets that are larger than those of \citet{mohri2024language,cherian2024large,rubintoles2025conformal}.

\section{E-SCORES}
\label{sec:e_scores}

We now describe the scores we propose to achieve \cref{eq:size_distortion}.
We defer the theoretical results that justify our design choices to \cref{subsec:theory-worst_and_e_values}; but intuitively, our scores must be reciprocals of the corresponding e-values.
Consequently, we call our proposed scores the \emph{e-scores}.

The functional form of our e-scores is influenced by \citet{gammerman1998learning}, who construct e-values for supervised learning under exchangeable data (used by \citet{balinsky2024enhancing,vovk2025conformal,gauthier2025backward} for the same).
Specifically, we define our e-score for each test response $\mathbf{y}^{n + 1} \in \mathds{Y} \rbr{g_{\pi} \rbr{x^{n + 1}}}$,
\begin{equation}
    \label{eq:e_score}
    \begin{gathered}
        s_{\text{e-score}} \rbr{x^{n + 1} , \mathbf{y}^{n + 1}} = \\
        \rbr{\frac{\rbr{n + 1} \cdot f \rbr{x^{n + 1} , \mathbf{y}^{n + 1}}}{f \rbr{x^{n + 1} , \mathbf{y}^{n + 1}} + \sum_{i = 1}^{n} f^{*} \rbr{x^{i} , \mathds{O} \rbr{x^{i} , g_{\pi} \rbr{x^{i}}}}}}^{- 1} ,
    \end{gathered}
\end{equation}
where $f$ is any function mapping a prompt $x$ and response $\mathbf{y}$ to a non-negative value $f \rbr{x , \mathbf{y}} \in \mathds{R}_{\geq 0}$, and,
\begin{equation*}
    f^{*} \rbr{x , \mathds{O} \rbr{x , g_{\pi} \rbr{x}}} = \max_{\rbr{\mathbf{y} , c} \in \mathds{O} \rbr{x , g_{\pi} \rbr{x}} : c = 0} f \rbr{x , \mathbf{y}} ,
\end{equation*}
is the maximum incorrect response value (set to 0 in the absence of an incorrect response).\footnote{We follow the convention $\nicefrac{a}{0} \! = \! 0$ if $a \! = \! 0$, otherwise $\pm \infty$.}
As a result, our e-scores compare a test response value with the incorrect calibration response values.
The specific definition of $f^{*}$ provides guarantees pertaining to the inclusion of any incorrect response (cf. \cref{subsec:theory-worst_and_e_values}), similar to the non-conformity functions in \citet{mohri2024language,cherian2024large,rubintoles2025conformal}.

For our e-scores to be measures of incorrectness, $f \rbr{x , \mathbf{y}}$ should intuitively be a proxy for the oracle $o \rbr{x , \mathbf{y}}$: high for correct responses and low for incorrect responses.
If the oracle were known, $f_{o} \rbr{x , \mathbf{y}}$ could be any monotonically increasing transformation of the oracle.
This includes, but is not limited to: (i) $o \rbr{x , \mathbf{y}}$, (ii) $\rbr{1 - o \rbr{x , \mathbf{y}}}^{- 1}$, and (iii) $o \rbr{x , \mathbf{y}} \cdot \rbr{1 - o \rbr{x , \mathbf{y}}}^{- 1}$.

However, since the oracle is unknown, we approximate it with $\hat{o}$.
Obtaining such an estimator is a binary classification problem (since the oracle is binary), where $\hat{o}$ predicts the probability of correctness, now in the range $\sbr{0 , 1}$.
Note that the data used for training $\hat{o}$ should be independent of the test and calibration data.
With an estimator $\hat{o}$, the above transformation options for $f_{\hat{o}} \rbr{x , \mathbf{y}}$ translate to e-scores with different ranges,
\begin{equation}
    \label{eq:f_options}
    \begin{gathered}
        f_{\hat{o}} \rbr{x , \mathbf{y}} = \\
        \begin{cases}
            \hat{o} \rbr{x , \mathbf{y}} \in \sbr{0 , 1} & \text{(for e-score 1)} \\
            \rbr{1 - \hat{o} \rbr{x , \mathbf{y}}}^{- 1} \in \sbr{1 , \infty} & \text{(for e-score 2)} \\
            \hat{o} \rbr{x , \mathbf{y}} \cdot \rbr{1 - \hat{o} \rbr{x , \mathbf{y}}}^{- 1} \in \sbr{0 , \infty} & \text{(for e-score 3)}
        \end{cases} .
    \end{gathered}
\end{equation}

Therefore, with an oracle estimator $\hat{o}$ and its transformation $f_{\hat{o}}$, we can compute our proposed e-scores in \cref{eq:e_score}.
We summarize this e-scoring mechanism in \cref{algo:e_score}.
In \cref{subsec:theory-worst_and_e_values}, we will show that our e-scores achieve \cref{eq:size_distortion} for any choice of $\hat{o}$ and $f_{\hat{o}}$, regardless of the possible errors in approximating $o$.

\begin{algorithm}[!t]
    \caption{E-Scores}
    \label{algo:e_score}
    \begin{small}
        \KwIn{Generative model $\pi$}
        \KwIn{Test prompt $x^{n + 1}$}
        \KwIn{Calibration prompt $x^{i}$ and labeled calibration responses $\mathds{O} \rbr{x^{i} , g_{\pi} \rbr{x^{i}}}$, for $i = 1 , \ldots , n$}
        \KwIn{Transformed oracle estimator $f_{\hat{o}}$}
        \KwOut{Test scored responses $\mathds{S} \rbr{x^{n + 1} , g_{\pi} \rbr{x^{n + 1}}}$}
        $g_{\pi} \rbr{x^{n + 1}} \gets \pi \rbr{\cdot | x^{n + 1}}$ \Comment*[r]{generation}
        $\mathds{S} \gets \emptyset$ \Comment*[r]{initialization}
        \For{$\mathbf{y}^{n + 1} \in \mathds{Y} \rbr{g_{\pi} \rbr{x^{n + 1}}}$ \Comment*[r]{response set}}{
            $v^{n + 1} \gets \rbr{\frac{\rbr{n + 1} \cdot f_{\hat{o}} \rbr{x^{n + 1} , \mathbf{y}^{n + 1}}}{f_{\hat{o}} \rbr{x^{n + 1} , \mathbf{y}^{n + 1}} + \sum_{i = 1}^{n} f_{\hat{o}}^{*} \rbr{x^{i} , \mathds{O} \rbr{x^{i} , g_{\pi} \rbr{x^{i}}}}}}^{- 1}$ \Comment*[r]{e-score}
            $\mathds{S} \gets \mathds{S} \cup \cbr{\rbr{\mathbf{y}^{n + 1} , v^{n + 1}}}$ \Comment*[r]{scored response}
        }
        \Return $\mathds{S}$
    \end{small}
\end{algorithm}

\subsection{Combining E-Scores}

With different options for the transformed oracle estimator in \cref{eq:f_options}, the use-case would determine the choice in practice.
Alternatively, without making assumptions about or restricting the use-cases, one can opt to combine multiple e-scores by first combining the underlying e-values.
We use the fact that simple averaging of e-values yields an admissible e-value \citep{vovk2021evalues}.
Let $s_{\text{e-score ($i$)}} \rbr{x^{n + 1} , \mathbf{y}^{n + 1}}$ for $i = 1 , 2 , 3$ be the three e-scores corresponding to the options in \cref{eq:f_options}.
Then, we can combine them into one e-score,
\begin{equation}
    \label{eq:combined_e_score}
    \begin{gathered}
        s_{\text{e-score (combined)}} \rbr{x^{n + 1} , \mathbf{y}^{n + 1}} = \\
        \rbr{\frac{\sum_{i = 1}^{3} \rbr{s_{\text{e-score ($i$)}} \rbr{x^{n + 1} , \mathbf{y}^{n + 1}}}^{- 1}}{3}}^{- 1} .
    \end{gathered}
\end{equation}
We will use this e-score by default, unless mentioned otherwise.
This is also used for the example in \cref{fig:example}.

\section{EXPERIMENTAL RESULTS}
\label{sec:experiments}

We now experimentally demonstrate the efficacy of our proposed e-scores with two practical settings.
We discuss the use-cases individually, then summarize the observed trends collectively.
We begin by stating the baselines, the metrics for comparisons, the time and memory complexities, and the experimental setup.
We also perform a worst-case analysis in \cref{appsec:experiments}, where the post-hoc $\alpha$'s maximize the size distortion from \cref{eq:size_distortion}.

\paragraph{Baselines}
We compare against p-value based scores, or \emph{p-scores}.
Analogous to our e-scores in \cref{eq:e_score}, the p-scores can be defined as the corresponding p-values,
\begin{equation}
    \label{eq:p_score}
    \begin{gathered}
        s_{\text{p-score}} \rbr{x^{n + 1} , \mathbf{y}^{n + 1}} = \\
        \frac{1 + \sum_{i = 1}^{n} \ind{\splitfrac{f \rbr{x^{n + 1} , \mathbf{y}^{n + 1}}}{\leq f^{*} \rbr{x^{i} , \mathds{O} \rbr{x^{i} , g_{\pi} \rbr{x^{i}}}}}}}{n + 1} ,
    \end{gathered}
\end{equation}
comparing a test response value with the incorrect calibration response values as relative ranks \citep{shafer2008tutorial,vovk2022algorithmic}.
\citet{mohri2024language,cherian2024large,rubintoles2025conformal} use such p-scores implicitly to achieve \cref{eq:size}, which we make explicit in \cref{appsec:related}.
Due to the reliance on relative ranks, the choice of the transformed oracle estimator in \cref{eq:f_options} does not matter, as they are monotonically increasing transformations of each other.

We also compare with the transformed oracle estimators in \cref{eq:f_options} directly, without any conversion to e- or p-scores.
These naive scores generally do not come with any statistical guarantees by themselves.
We will use these scores for our worst-case analysis in \cref{appsec:experiments}.

\paragraph{Metrics}
Our comparisons are based on the following.
\begin{itemize}
    \item
    \textit{Size distortion.}
    This is the most important metric, from our desideratum in \cref{eq:size_distortion}.
    We report its empirical mean, which is desired to be at most 1.
    
    \item
    \textit{Error vs. $\alpha$.}
    While we aim to control size distortion, the expected error to $\alpha$ ratio, one might also be interested in the expected error and expected $\alpha$ individually.
    We report empirical means for both.
    We ideally want the observed error to be lower than the tolerance level (mean error $\leq$ mean $\alpha$).
    
    \item
    \textit{Precision vs. recall.}
    We provide guarantees on the inclusion of any incorrect response.
    Simultaneously, we do not wish to exclude many correct responses.
    As a result, we report the empirical means for precision (fraction of correct responses among those included) and recall (fraction of correct responses included).
    We compare using the precision-recall curves (higher is better for both).
\end{itemize}

\paragraph{Memory and Time Complexities}
Our e-scores are cheaper to compute than the p-scores, in memory and in time.
For a given test prompt-response pair, p-scores compute relative ranks with the calibration data (cf. \cref{eq:p_score}).
This requires memory and time that grows linearly in the amount of calibration data $n$ for every individual test prompt-response pair.
Conversely, our e-scores compute a sum over the calibration data (cf. \cref{eq:e_score}).
This requires constant memory and time that grows linearly in $n$.
Furthermore, this is a one time cost, amortized over all test prompt-response pairs.

\paragraph{Setup}
We randomly split the data 50-50\% into the test and calibration data (no training data is required as we use pre-trained oracle estimators).
The metrics are averaged over 100 such random splits.
We use an NVIDIA A100 GPU for the pre-trained oracle estimators; the remaining computations run on a CPU.

\subsection{Mathematical Factuality}
\label{subsec:experiments-processbench}

\begin{figure*}[!t]
    \begin{center}
        \begin{subfigure}{\linewidth}
            \includegraphics[width=\linewidth]{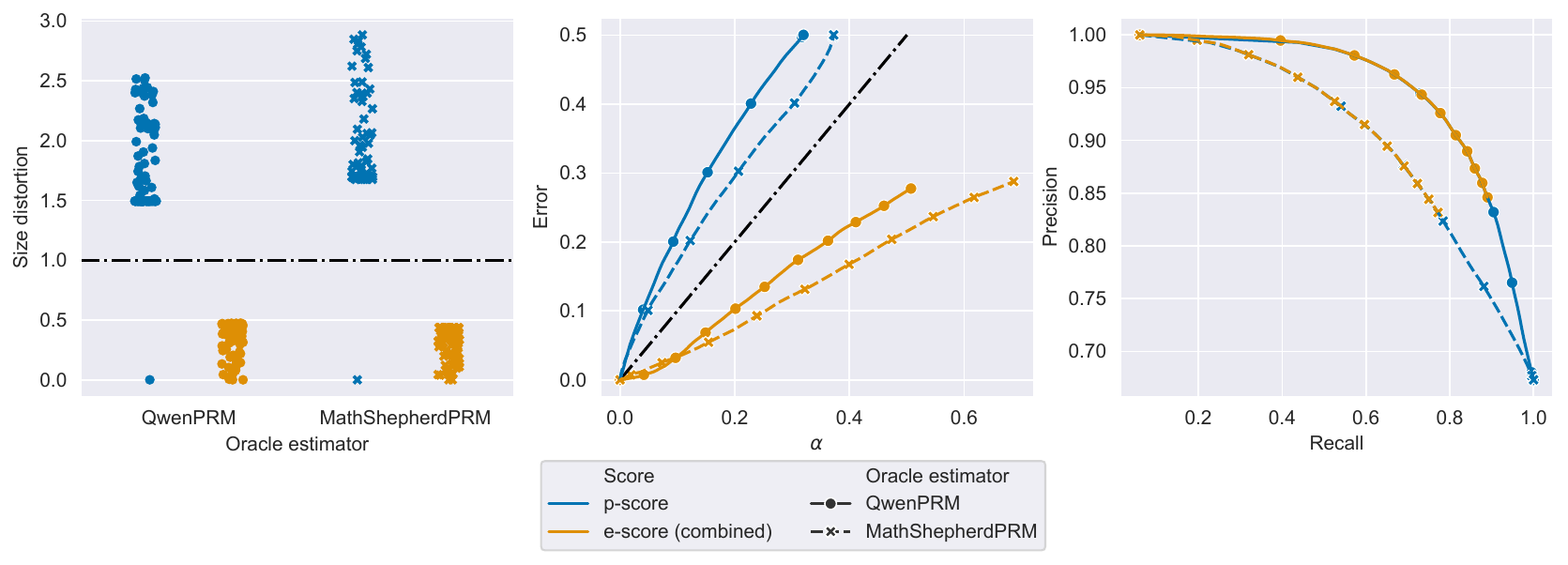}
            \caption{
                \textbf{Max-constrained adaptive $\pmb{\alpha}$ strategies.}
                We set $\alpha_{\max} = 0 , .01 , .02 , \ldots , .99 , 1$ (cf. \cref{subsec:alpha_strategies}).
            }
            \label{fig:processbench-alphamax}
            \vspace{0.5\abovecaptionskip}
        \end{subfigure}
        \begin{subfigure}{\linewidth}
            \vspace{0.5\abovecaptionskip}
            \includegraphics[width=\linewidth]{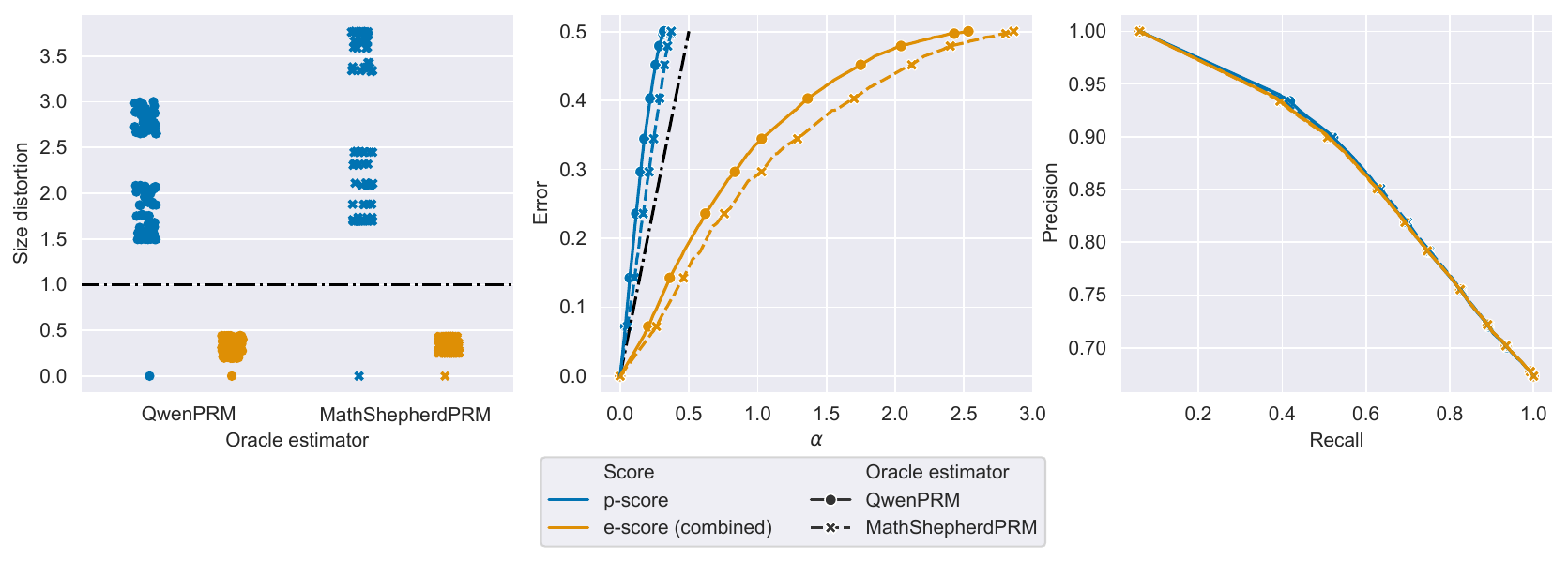}
            \caption{
                \textbf{Fractional inclusion strategies.}
                We set $\lambda = 0 , .01 , .02 , \ldots , .99 , 1$ (cf. \cref{subsec:alpha_strategies}).
            }
            \label{fig:processbench-fracinclusion}
        \end{subfigure}
    \end{center}
    \caption{
        \textbf{Scores for mathematical factuality.}
        We use the setting in \cref{subsec:experiments-processbench} to compare our proposed e-scores (in orange) against p-scores (in blue).
        The left graphs plot size distortion (cf. \cref{eq:size_distortion}).
        The center graphs plot mean error vs. mean $\alpha$ (where the dashed black line is the identity line).
        The right graphs plot mean precision vs. mean recall (with the e-scores curves overlapping and hiding part of or the entire p-scores curves).
    }
    \label{fig:processbench}
\end{figure*}

ProcessBench \citep{zheng2025processbench} is a mathematical reasoning benchmark.
It contains prompts from GSM8K \citep{cobbe2021training}, MATH \citep{hendrycks2021measuring}, OlympiadBench \citep{he2024olympiadbench}, and Omni-MATH \citep{gao2025omnimath}.
The responses are first generated by 12 open-source LLMs \citep{llama2024llama3,yang2024qwen2,yang2024qwen2.5math,qwen2025qwen2.5}, then separated into multiple steps/sub-responses using Qwen2.5-72B-Instruct \citep{qwen2025qwen2.5}.
Lastly, human experts annotate the earliest-occurring incorrect sub-response.

Let $\mathbf{y} = \rbr{y_{1} , y_{2} , \ldots}$ be a generated response and $i$ be its annotation.
Then, the responses $\mathbf{y}_{\leq j}$ for $j = i , \ldots , \nbr{\mathbf{y}}$ are deemed incorrect as they contain the incorrect $i$-th sub-response, whereas the responses for $j = 1 , \ldots , i - 1$ are correct.
\cref{fig:example} illustrates one such example.
This determines the correctness or factuality of the responses.

\citet{zheng2025processbench} also benchmark math-based process reward models that predict the correctness probabilities of sub-responses individually.
However, we want an oracle estimator to predict the correctness probability of (partial) responses.
We follow \citet{lightman2024lets} to obtain this by multiplying the individual (conditional) sub-response predictions like conditional probabilities,
\begin{equation*}
    \hat{o} \rbr{x , \mathbf{y}} = \prod_{i = 1}^{\nbr{\mathbf{y}}} \hat{o} \rbr{x , y_{i} \mid \mathbf{y}_{< i}} .
\end{equation*}
We consider two process reward models for such oracle estimators: (i) Qwen2.5-Math-7B-PRM800K (or QwenPRM) \citep{zheng2025processbench} and (ii) Math-Shepherd-PRM-7B (or MathShepherdPRM) \citep{wang2024mathshepherd}.

We examine both the post-hoc $\alpha$ strategies described in \cref{subsec:alpha_strategies}, and illustrate the results in \cref{fig:processbench}.
Furthermore, the example in \cref{fig:example} is from this use-case.

\subsection{Property Constraints Satisfaction}
\label{subsec:experiments-ultrafeedback}

\begin{figure*}[!t]
    \begin{center}
        \begin{subfigure}{\linewidth}
            \includegraphics[width=\linewidth]{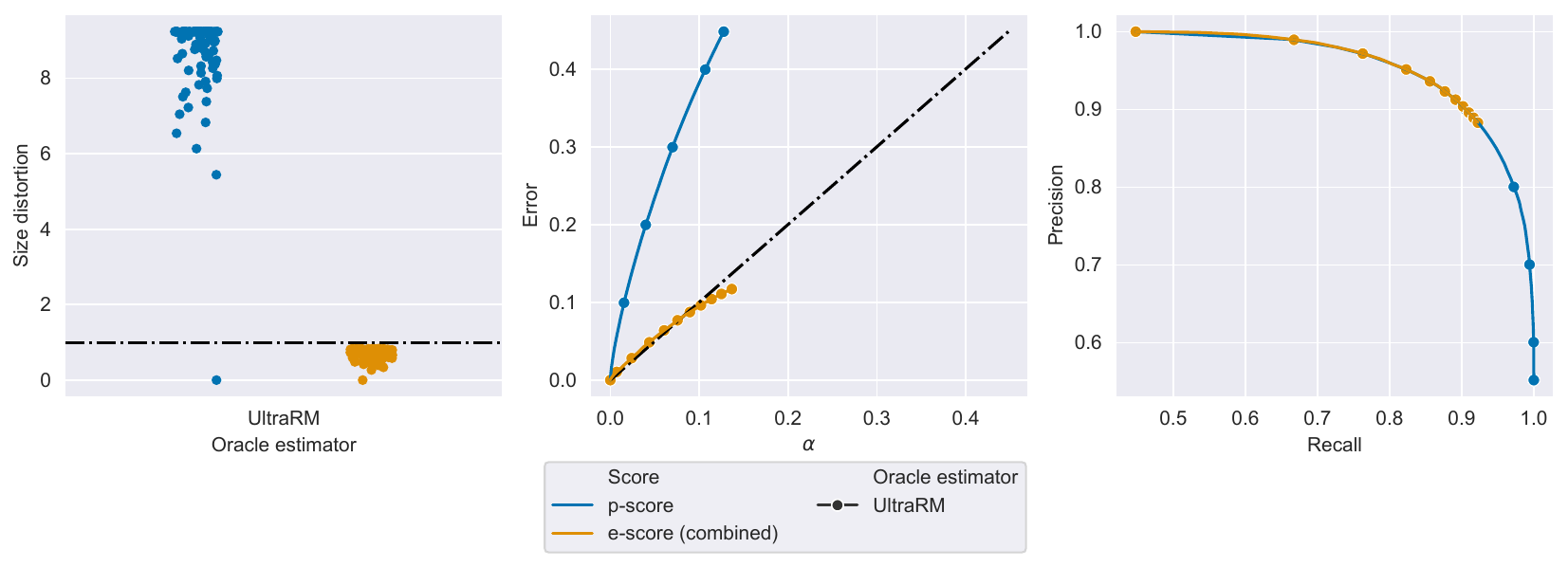}
            \caption{
                \textbf{Helpfulness and instruction-following.}
                A response is correct if its helpfulness and instruction-following ratings $\geq 4$.
            }
            \label{fig:ultrafeedback-evolinstruct-alphamax}
            \vspace{0.5\abovecaptionskip}
        \end{subfigure}
        \begin{subfigure}{\linewidth}
            \vspace{0.5\abovecaptionskip}
            \includegraphics[width=\linewidth]{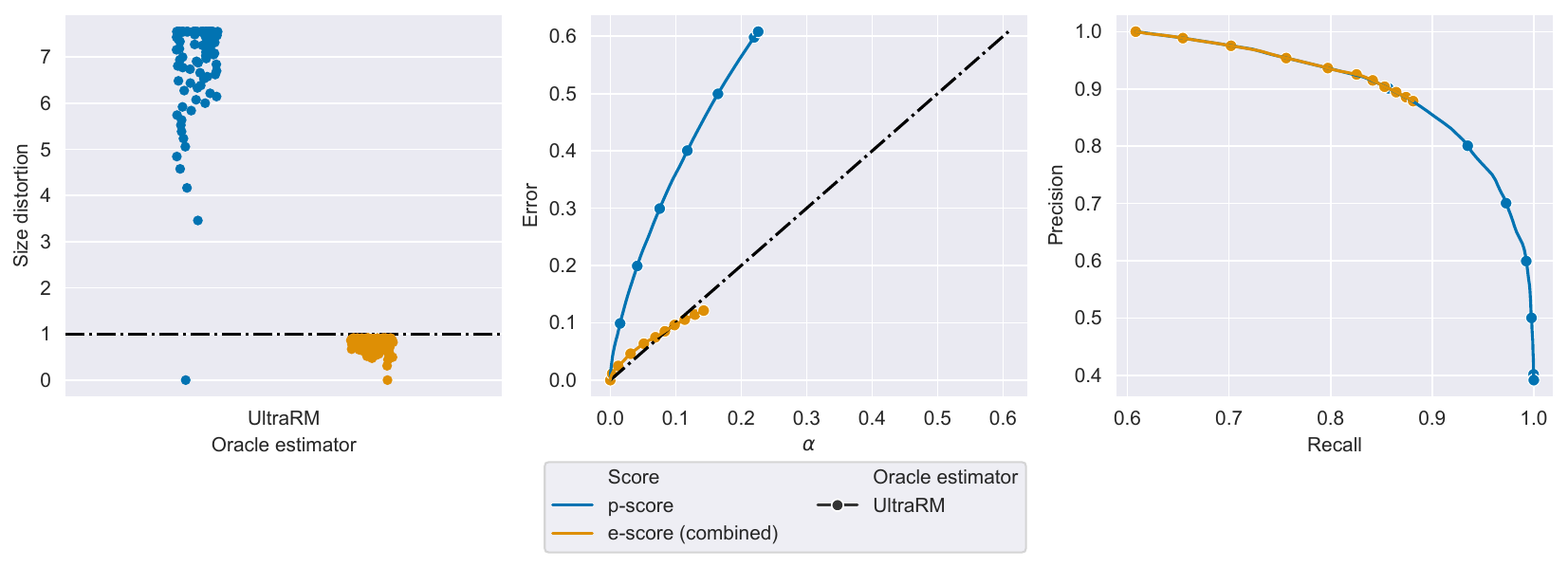}
            \caption{
                \textbf{Honesty and truthfulness.}
                A response is correct if its honesty and truthfulness ratings $= 5$.
            }
            \label{fig:ultrafeedback-truthfulqa-alphamax}
        \end{subfigure}
    \end{center}
    \caption{
        \textbf{Scores for property constraints satisfaction.}
        We use the setting in \cref{subsec:experiments-ultrafeedback} to compare our proposed e-scores (in orange) against p-scores (in blue).
        We consider different max-constrained adaptive $\alpha$ strategies, setting $\alpha_{\max} = 0 , .01 , .02 , \ldots , .99 , 1$ (cf. \cref{subsec:alpha_strategies}).
        The left graphs plot size distortion (cf. \cref{eq:size_distortion}).
        The center graphs plot mean error vs. mean $\alpha$ (where the dashed black line is the identity line).
        The right graphs plot mean precision vs. mean recall (with the e-scores curves overlapping and hiding part of the p-scores curves).
    }
\end{figure*}

UltraFeedback \citep{cui2024ultrafeedback} is a diverse and fine-grained preference dataset.
It contains prompts from 6 benchmarks (we will only use Evol-Instruct \citep{xu2024wizardlm} and TruthfulQA \citep{lin2022truthfulqa}) and responses from 17 commercial and open-source LLMs \citep{chiang2023vicuna,tunstall2023creating,taori2023stanford,touvron2023llama2,biderman2023pythia,almazrouei2023falcon,ding2023enhancing,openai2024gpt4,xu2024wizardlm}.
It also employs GPT-4 \citep{openai2024gpt4} to provide a rating for every response on four different criteria: helpfulness, honesty, instruction-following, and truthfulness; these are numerical ratings from 1 to 5.

In practice, users are often interested in responses that satisfy certain desirable properties; this is equivalent to thresholding or constraining the property ratings to their corresponding desirable values \citep{dhillon2025l3ms}.
We use such a constraining mechanism to define the correctness of responses for two different use-cases.

\paragraph{Helpfulness and Instruction-Following}
We define a response to be correct if both its helpfulness and instruction-following ratings are either 4 or higher.
We use prompts from the Evol-Instruct benchmark \citep{xu2024wizardlm}.
We illustrate these results in \cref{fig:ultrafeedback-evolinstruct-alphamax}.

\paragraph{Honesty and Truthfulness}
We define a response to be correct if both its honesty and truthfulness ratings are 5.
We use prompts from the TruthfulQA benchmark \citep{lin2022truthfulqa}.
We illustrate these results in \cref{fig:ultrafeedback-truthfulqa-alphamax}.

\citet{cui2024ultrafeedback} also provide a reward model, UltraRM, that predicts a real-valued preference for a response.
We follow the Bradley-Terry model \citep{bradley1952rank} (although without a reference preference value, or equivalently, a reference preference of 0) and append the pre-trained UltraRM with a sigmoid operator, updating its range to $\sbr{0 , 1}$.
We use this as our oracle estimator.

Note that these use-cases consider the full responses with $\nbr{\mathbf{y}} = 1$.
As a result, we only examine the max-constrained adaptive $\alpha$ strategy (cf. \cref{subsec:alpha_strategies}); fractional inclusion would include or exclude all responses.

\subsection{Observed Trends}
\label{subsec:experiments-trends}

The trends we observe for the different experimental use-cases are consistent; we summarize them together.

\paragraph{Size Distortion}
Our proposed e-scores reliably upper bound size distortion to 1 and satisfy \cref{eq:size_distortion}, corroborating our theory in \cref{subsec:theory-worst_and_e_values}.
Conversely, p-scores are unable to achieve this; the only time they experimentally do is when all responses (correct and incorrect) are excluded, achieving 0 error by default.
    
\paragraph{Error vs. $\pmb{\alpha}$}
Our proposed e-scores consistently obtain a mean error lower than or approximately equal to the mean tolerance $\alpha$.
Conversely, p-scores consistently obtain a mean error higher than the mean tolerance $\alpha$.

\paragraph{Precision vs. Recall}
The precision-recall curves of our proposed e-scores overlap with those of the p-scores.
In satisfying \cref{eq:size_distortion}, the e-scores are more conservative and prefer maintaining high precision over high recall.
Consequently, restricting $\alpha$'s to be $\leq 1$ (under the max-constrained adaptive $\alpha$ strategies) restricts the e-score recalls compared to the p-score recalls, resulting in partial overlap.
However, removing this restriction (under the fractional inclusion strategies) retains complete overlap of the e- and p-score precision-recall curves.

\paragraph{Oracle Estimator}
The choice of the oracle estimator impacts the metrics.
This is best illustrated by the precision-recall curves in \cref{fig:processbench-alphamax}, where QwenPRM achieves higher precisions and recalls compared to MathShepherdPRM.
This is expected as the former is comparatively more accurate \citep{zheng2025processbench}.

\section{THEORETICAL RESULTS}
\label{sec:theory}

We present our theoretical results here.
We generalize the response set from \cref{eq:simple_response_set} to a larger set, and show that our e-scores satisfy our desideratum in \cref{eq:size_distortion}.

\subsection{Super-Set of Responses}
\label{subsec:theory-setup}

We intend to make the response set from \cref{eq:simple_response_set} as large as possible, while maintaining statistical guarantees.
This would enable the assessment of a larger set of responses, opening avenues for more diverse applications and use-cases.
So, we consider the responses,
\begin{equation}
    \label{eq:response_set}
    \mathds{Y} \rbr{g_{\pi} \rbr{x}} = \cup_{\sigma} \cbr{\mathbf{y}_{\leq i} : \mathbf{y} = \sigma \rbr{g_{\pi} \rbr{x}} , i = 1 , \ldots , \nbr{\mathbf{y}}} ,
\end{equation}
where $\sigma \rbr{g_{\pi} \rbr{x}}$ is a permuted version of the generated response $g_{\pi} \rbr{x}$, and the union is over all permutations.

If we fix $\sigma$ to the identity ordering only, we recover \cref{eq:simple_response_set}.
Similarly, \citet{rubintoles2025conformal} restrict $\sigma$ to orderings (what they call topological orderings of an approximate deducibility graph) obtained from GPT-4o \citep{openai2024gpt4o}.
Lastly, \citet{mohri2024language,cherian2024large} do not account for the inherent ordering of the sub-responses to make up a response.
Therefore, \cref{eq:response_set} is a super-set of responses, containing all the response sets discussed above.
In fact, we believe that it is the largest set of responses to consider when given a \emph{single} generated response $g_{\pi} \rbr{x}$.

Instead of the full response set in \cref{eq:response_set}, one might choose to use a sub-set depending on the use-case, while maintaining guarantees.
For example, we use \cref{eq:simple_response_set} in \cref{subsec:experiments-processbench} as it is tailor-made for that benchmark.

\subsection{Worst-Case Analysis and E-Values}
\label{subsec:theory-worst_and_e_values}

We are interested in achieving the desideratum in \cref{eq:size_distortion} for any post-hoc $\alpha$ that a user might choose.
Without restricting the user's choice, we will analyze the setting where $\alpha \rbr{x , \mathds{S} \rbr{x , g_{\pi} \rbr{x}}}$ maximizes size distortion.
If \cref{eq:size_distortion} is satisfied under this worst-case setting, it will also be satisfied under any post-hoc $\alpha$.

Note that a response is included in the filtered set $\mathds{S}_{\alpha} \rbr{x , g_{\pi} \rbr{x}}$ if and only if its score is $\leq \alpha$.
As a result, we can re-write the inclusion of an incorrect response as at least one incorrect response having a score $\leq \alpha$,
\begin{equation*}
    \begin{gathered}
        \exists \rbr{\mathbf{y} , \cdot} \in \mathds{S}_{\alpha} \rbr{x , g_{\pi} \rbr{x}} \text{ s.t. } o \rbr{x , \mathbf{y}} = 0 \\
        \iff \min_{\rbr{\mathbf{y} , c} \in \mathds{O} \rbr{x , g_{\pi} \rbr{x}} : c = 0} s \rbr{x , \mathbf{y}} \leq \alpha .
    \end{gathered}
\end{equation*}
Therefore, the worst-case size distortion simplifies to,
\begin{equation*}
    \begin{gathered}
        \ev{\max_{\alpha \in \mathds{R}_{\geq 0}} \frac{\ind{\min_{\rbr{\mathbf{Y} , C} \in \mathds{O} \rbr{X , g_{\pi} \rbr{X}} : C = 0} s \rbr{X , \mathbf{Y}} \leq \alpha}}{\alpha}} \\
        = \ev{\rbr{\min_{\rbr{\mathbf{Y} , C} \in \mathds{O} \rbr{X , g_{\pi} \rbr{X}} : C = 0} s \rbr{X , \mathbf{Y}}}^{- 1}} ,
    \end{gathered}
\end{equation*}
because $\alpha$ is set to the smallest value for which the indicator evaluates to 1, otherwise the whole term is 0.

To upper bound the above expectation by 1 is equivalent to requiring $\rbr{\min_{\rbr{\mathbf{y} , c} \in \mathds{O} \rbr{x , g_{\pi} \rbr{x}} : c = 0} s \rbr{x , \mathbf{y}}}^{- 1}$ to be an e-value (by definition), and hence the specific choice of our proposed e-scores in \cref{eq:e_score}.
Indeed, if we use our e-scores here, the above term simplifies to,
\begin{equation*}
    \begin{gathered}
        \rbr{\min_{\rbr{\mathbf{y}^{n + 1} , c^{n + 1}} \in \mathds{O} \rbr{x^{n + 1} , g_{\pi} \rbr{x^{n + 1}}} : c^{n + 1} = 0} s \rbr{x^{n + 1} , \mathbf{y}^{n + 1}}}^{- 1} \\
        = \frac{\rbr{n + 1} \cdot f^{*} \rbr{x^{n + 1} , \mathds{O} \rbr{x^{n + 1} , g_{\pi} \rbr{x^{n + 1}}}}}{\sum_{i = 1}^{n + 1} f^{*} \rbr{x^{i} , \mathds{O} \rbr{x^{i} , g_{\pi} \rbr{x^{i}}}}} ,
    \end{gathered}
\end{equation*}
which is an e-value under exchangeable data for any non-negative function $f$ \citep{gammerman1998learning}.

Lastly, since our e-scores satisfy \cref{eq:size_distortion} under this worst-case setting, they will satisfy \cref{eq:size_distortion} under any post-hoc $\alpha$.
We summarize this theoretical result below, and provide the detailed derivation in \cref{appsec:theory}.

\begin{restatable}{theorem}{theoremsizedistortion}
\label{theorem:size_distortion}
If the test and the calibration prompts are exchangeable, then, our proposed e-scores in \cref{eq:e_score,eq:combined_e_score} upper bound the size distortion (marginal over the test and the calibration prompts) by 1, as in \cref{eq:size_distortion}.
\end{restatable}

\section{CONCLUSIONS}
\label{sec:conclusions}

In this paper, we studied the problem of achieving statistical guarantees for a post-hoc notion of error, namely size distortion, for generative model outputs.
We proposed e-scores, based on e-values, as measures of incorrectness.
We proved theoretically that our proposed e-scores achieve the desired post-hoc guarantees, which we corroborated with experimental results.
In doing so, e-scores provide users the flexibility of choosing data-dependent tolerance levels $\alpha$.
We also showed that our guarantees extend to a large set of responses, opening up possibilities for more diverse applications.

\paragraph{Future Work}
Our experiments demonstrated that the choice of the oracle estimator impacts metrics such as the precision-recall curves.
While we used pre-trained estimators, they could be trained for specific applications to strengthen the e-scores.
Furthermore, while we considered size distortion as our post-hoc notion of error, other candidates exist, although they are not well understood.
\citet{koning2024posthoc} discusses some alternatives, which could be investigated in the future.

\bibliographystyle{abbrvnat}
\bibliography{main}

@misc{almazrouei2023falcon,
    title={The {F}alcon Series of Open Language Models},
    author={Ebtesam Almazrouei and Hamza Alobeidli and Abdulaziz Alshamsi and Alessandro Cappelli and Ruxandra Cojocaru and M\'{e}rouane Debbah and \'{E}tienne Goffinet and Daniel Hesslow and Julien Launay and Quentin Malartic and Daniele Mazzotta and Badreddine Noune and Baptiste Pannier and Guilherme Penedo},
    year={2023},
    eprint={2311.16867},
    archivePrefix={arXiv},
    primaryClass={cs.CL},
    url={https://arxiv.org/abs/2311.16867}
}

@inproceedings{balinsky2024enhancing,
    title={Enhancing Conformal Prediction Using E-Test Statistics},
    author={Balinsky, Alexander A. and Balinsky, Alexander David},
    booktitle={Proceedings of the Thirteenth Symposium on Conformal and Probabilistic Prediction with Applications},
    pages={65--72},
    year={2024},
    editor={Vantini, Simone and Fontana, Matteo and Solari, Aldo and Bostr\"{o}m, Henrik and Carlsson, Lars},
    volume={230},
    series={Proceedings of Machine Learning Research},
    month={09--11 Sep},
    publisher={PMLR},
    url={https://proceedings.mlr.press/v230/balinsky24a.html}
}

@inproceedings{biderman2023pythia,
    title={{P}ythia: A Suite for Analyzing Large Language Models Across Training and Scaling},
    author={Biderman, Stella and Schoelkopf, Hailey and Anthony, Quentin Gregory and Bradley, Herbie and O'Brien, Kyle and Hallahan, Eric and Khan, Mohammad Aflah and Purohit, Shivanshu and Prashanth, Usvsn Sai and Raff, Edward and Skowron, Aviya and Sutawika, Lintang and Van Der Wal, Oskar},
    booktitle={Proceedings of the 40th International Conference on Machine Learning},
    pages={2397--2430},
    year={2023},
    editor={Krause, Andreas and Brunskill, Emma and Cho, Kyunghyun and Engelhardt, Barbara and Sabato, Sivan and Scarlett, Jonathan},
    volume={202},
    series={Proceedings of Machine Learning Research},
    month={23--29 Jul},
    publisher={PMLR},
    url={https://proceedings.mlr.press/v202/biderman23a.html}
}

@article{bradley1952rank,
    author={Bradley, Ralph Allan and Terry, Milton E.},
    title={Rank Analysis of Incomplete Block Designs: The Method of Paired Comparisons},
    journal={Biometrika},
    volume={39},
    number={3--4},
    pages={324--345},
    year={1952},
    month={12},
    issn={0006-3444},
    doi={10.1093/biomet/39.3-4.324},
    url={https://doi.org/10.1093/biomet/39.3-4.324},
    eprint={https://academic.oup.com/biomet/article-pdf/39/3-4/324/930466/39-3-4-324.pdf}
}

@misc{carney2016position,
    author={Carney, Dana R.},
    title={My position on ``{P}ower {P}oses''},
    year={2016},
    url={https://faculty.haas.berkeley.edu/dana_carney/pdf_my\%20position\%20on\%20power\%20poses.pdf}
}

@inproceedings{cherian2024large,
    author={Cherian, John J. and Gibbs, Isaac and Cand\`{e}s, Emmanuel J.},
    booktitle={Advances in Neural Information Processing Systems},
    editor={A. Globerson and L. Mackey and D. Belgrave and A. Fan and U. Paquet and J. Tomczak and C. Zhang},
    pages={114812--114842},
    publisher={Curran Associates, Inc.},
    title={Large language model validity via enhanced conformal prediction methods},
    url={https://proceedings.neurips.cc/paper_files/paper/2024/file/d02ff1aeaa5c268dc34790dd1ad21526-Paper-Conference.pdf},
    volume={37},
    year={2024}
}

@misc{chiang2023vicuna,
    title={{V}icuna: An Open-Source Chatbot Impressing {GPT}-4 with 90\%* {C}hat{GPT} Quality},
    url={https://lmsys.org/blog/2023-03-30-vicuna/},
    author={Chiang, Wei-Lin and Li, Zhuohan and Lin, Zi and Sheng, Ying and Wu, Zhanghao and Zhang, Hao and Zheng, Lianmin and Zhuang, Siyuan and Zhuang, Yonghao and Gonzalez, Joseph E. and Stoica, Ion and Xing, Eric P.},
    month={March},
    year={2023}
}

@misc{cobbe2021training,
    title={Training Verifiers to Solve Math Word Problems},
    author={Karl Cobbe and Vineet Kosaraju and Mohammad Bavarian and Mark Chen and Heewoo Jun and Lukasz Kaiser and Matthias Plappert and Jerry Tworek and Jacob Hilton and Reiichiro Nakano and Christopher Hesse and John Schulman},
    year={2021},
    eprint={2110.14168},
    archivePrefix={arXiv},
    primaryClass={cs.LG},
    url={https://arxiv.org/abs/2110.14168}
}

@inproceedings{cui2024ultrafeedback,
    title={{U}ltra{F}eedback: Boosting Language Models with Scaled {AI} Feedback},
    author={Cui, Ganqu and Yuan, Lifan and Ding, Ning and Yao, Guanming and He, Bingxiang and Zhu, Wei and Ni, Yuan and Xie, Guotong and Xie, Ruobing and Lin, Yankai and Liu, Zhiyuan and Sun, Maosong},
    booktitle={Proceedings of the 41st International Conference on Machine Learning},
    pages={9722--9744},
    year={2024},
    editor={Salakhutdinov, Ruslan and Kolter, Zico and Heller, Katherine and Weller, Adrian and Oliver, Nuria and Scarlett, Jonathan and Berkenkamp, Felix},
    volume={235},
    series={Proceedings of Machine Learning Research},
    month={21--27 Jul},
    publisher={PMLR},
    url={https://proceedings.mlr.press/v235/cui24f.html}
}

@inproceedings{dhillon2025l3ms,
    author={Dhillon, Guneet S. and Shi, Xingjian and Teh, Yee Whye and Smola, Alex},
    booktitle={International Conference on Representation Learning},
    editor={Yue, Y. and Garg, A. and Peng, N. and Sha, F. and Yu, R.},
    pages={58300--58314},
    title={{L3M}s --- {L}agrange Large Language Models},
    url={https://proceedings.iclr.cc/paper_files/paper/2025/file/92d3d2a9801211ca3693ccb2faa1316f-Paper-Conference.pdf},
    volume={2025},
    year={2025}
}

@inproceedings{ding2023enhancing,
    title={Enhancing Chat Language Models by Scaling High-quality Instructional Conversations},
    author={Ding, Ning and Chen, Yulin and Xu, Bokai and Qin, Yujia and Hu, Shengding and Liu, Zhiyuan and Sun, Maosong and Zhou, Bowen},
    editor={Bouamor, Houda and Pino, Juan and Bali, Kalika},
    booktitle={Proceedings of the 2023 Conference on Empirical Methods in Natural Language Processing},
    month=dec,
    year={2023},
    address={Singapore},
    publisher={Association for Computational Linguistics},
    url={https://aclanthology.org/2023.emnlp-main.183/},
    doi={10.18653/v1/2023.emnlp-main.183},
    pages={3029--3051}
}

@inproceedings{gammerman1998learning,
    title={Learning by Transduction},
    author={Alexander Gammerman and Volodya Vovk and Vladimir Vapnik},
    booktitle={Proceedings of the Fourteenth Conference on Uncertainty in Artificial Intelligence},
    pages={148--155},
    year={1998},
    editor={Gregory F. Cooper and Seraf\'{\i}n Moral},
    month={24--26 July},
    publisher={Morgan Kaufmann Publishers Inc.}
}

@inproceedings{gao2025omnimath,
    title={{O}mni-{MATH}: A Universal Olympiad Level Mathematic Benchmark for Large Language Models},
    author={Bofei Gao and Feifan Song and Zhe Yang and Zefan Cai and Yibo Miao and Qingxiu Dong and Lei Li and Chenghao Ma and Liang Chen and Runxin Xu and Zhengyang Tang and Benyou Wang and Daoguang Zan and Shanghaoran Quan and Ge Zhang and Lei Sha and Yichang Zhang and Xuancheng Ren and Tianyu Liu and Baobao Chang},
    booktitle={The Thirteenth International Conference on Learning Representations},
    year={2025},
    url={https://openreview.net/forum?id=yaqPf0KAlN}
}

@inproceedings{gauthier2025backward,
    title={Backward Conformal Prediction},
    author={Etienne Gauthier and Francis Bach and Michael I. Jordan},
    booktitle={The Thirty-ninth Annual Conference on Neural Information Processing Systems},
    year={2025},
    url={https://openreview.net/forum?id=23ichdd74N}
}

@misc{gemini2025gemini2.5,
    title={{G}emini 2.5: Pushing the Frontier with Advanced Reasoning, Multimodality, Long Context, and Next Generation Agentic Capabilities},
    author={{Gemini Team}},
    year={2025},
    eprint={2507.06261},
    archivePrefix={arXiv},
    primaryClass={cs.CL},
    url={https://arxiv.org/abs/2507.06261}
}

@article{grunwald2024beyond,
    author={Peter D. Gr\"{u}nwald},
    title={Beyond {N}eyman–{P}earson: E-values enable hypothesis testing with a data-driven alpha},
    journal={Proceedings of the National Academy of Sciences},
    volume={121},
    number={39},
    pages={e2302098121},
    year={2024},
    doi={10.1073/pnas.2302098121},
    URL={https://www.pnas.org/doi/abs/10.1073/pnas.2302098121},
    eprint={https://www.pnas.org/doi/pdf/10.1073/pnas.2302098121}
}

@article{grunwald2024safe,
    author={Gr\"{u}nwald, Peter and de Heide, Rianne and Koolen, Wouter},
    title={Safe testing},
    journal={Journal of the Royal Statistical Society Series B: Statistical Methodology},
    volume={86},
    number={5},
    pages={1091--1128},
    year={2024},
    month={03},
    issn={1369-7412},
    doi={10.1093/jrsssb/qkae011},
    url={https://doi.org/10.1093/jrsssb/qkae011},
    eprint={https://academic.oup.com/jrsssb/article-pdf/86/5/1091/60648648/qkae011.pdf}
}

@article{gupta2022nested,
    title={Nested conformal prediction and quantile out-of-bag ensemble methods},
    journal={Pattern Recognition},
    volume={127},
    pages={108496},
    year={2022},
    issn={0031-3203},
    doi={https://doi.org/10.1016/j.patcog.2021.108496},
    url={https://www.sciencedirect.com/science/article/pii/S0031320321006725},
    author={Chirag Gupta and Arun K. Kuchibhotla and Aaditya Ramdas}
}

@inproceedings{he2024olympiadbench,
    title={{O}lympiad{B}ench: A Challenging Benchmark for Promoting {AGI} with Olympiad-Level Bilingual Multimodal Scientific Problems},
    author={He, Chaoqun and Luo, Renjie and Bai, Yuzhuo and Hu, Shengding and Thai, Zhen and Shen, Junhao and Hu, Jinyi and Han, Xu and Huang, Yujie and Zhang, Yuxiang and Liu, Jie and Qi, Lei and Liu, Zhiyuan and Sun, Maosong},
    editor={Ku, Lun-Wei and Martins, Andre and Srikumar, Vivek},
    booktitle={Proceedings of the 62nd Annual Meeting of the Association for Computational Linguistics (Volume 1: Long Papers)},
    month=aug,
    year={2024},
    address={Bangkok, Thailand},
    publisher={Association for Computational Linguistics},
    url={https://aclanthology.org/2024.acl-long.211/},
    doi={10.18653/v1/2024.acl-long.211},
    pages={3828--3850}
}

@inproceedings{hendrycks2021measuring,
    author={Hendrycks, Dan and Burns, Collin and Kadavath, Saurav and Arora, Akul and Basart, Steven and Tang, Eric and Song, Dawn and Steinhardt, Jacob},
    booktitle={Proceedings of the Neural Information Processing Systems Track on Datasets and Benchmarks},
    editor={J. Vanschoren and S. Yeung},
    pages={},
    title={Measuring Mathematical Problem Solving With the {MATH} Dataset},
    url={https://datasets-benchmarks-proceedings.neurips.cc/paper_files/paper/2021/file/be83ab3ecd0db773eb2dc1b0a17836a1-Paper-round2.pdf},
    volume={1},
    year={2021}
}

@article{huang2025survey,
    author={Huang, Lei and Yu, Weijiang and Ma, Weitao and Zhong, Weihong and Feng, Zhangyin and Wang, Haotian and Chen, Qianglong and Peng, Weihua and Feng, Xiaocheng and Qin, Bing and Liu, Ting},
    title={A Survey on Hallucination in Large Language Models: Principles, Taxonomy, Challenges, and Open Questions},
    year={2025},
    issue_date={March 2025},
    publisher={Association for Computing Machinery},
    address={New York, NY, USA},
    volume={43},
    number={2},
    issn={1046-8188},
    url={https://doi.org/10.1145/3703155},
    doi={10.1145/3703155},
    journal={ACM Transactions on Information Systems},
    month=jan,
    articleno={42},
    numpages={55}
}

@misc{koning2024posthoc,
    title={Post-hoc $\alpha$ Hypothesis Testing and the Post-hoc $p$-value},
    author={Nick W. Koning},
    year={2024},
    eprint={2312.08040},
    archivePrefix={arXiv},
    primaryClass={math.ST},
    url={https://arxiv.org/abs/2312.08040}
}

@inproceedings{lightman2024lets,
    title={Let's Verify Step by Step},
    author={Hunter Lightman and Vineet Kosaraju and Yuri Burda and Harrison Edwards and Bowen Baker and Teddy Lee and Jan Leike and John Schulman and Ilya Sutskever and Karl Cobbe},
    booktitle={The Twelfth International Conference on Learning Representations},
    year={2024},
    url={https://openreview.net/forum?id=v8L0pN6EOi}
}

@inproceedings{lin2022truthfulqa,
    title={{T}ruthful{QA}: Measuring How Models Mimic Human Falsehoods},
    author={Lin, Stephanie and Hilton, Jacob and Evans, Owain},
    editor={Muresan, Smaranda and Nakov, Preslav and Villavicencio, Aline},
    booktitle={Proceedings of the 60th Annual Meeting of the Association for Computational Linguistics (Volume 1: Long Papers)},
    month=may,
    year={2022},
    address={Dublin, Ireland},
    publisher={Association for Computational Linguistics},
    url={https://aclanthology.org/2022.acl-long.229/},
    doi={10.18653/v1/2022.acl-long.229},
    pages={3214--3252}
}

@misc{llama2024llama3,
    title={The {LL}a{MA} 3 Herd of Models},
    author={{LLaMA Team}},
    year={2024},
    eprint={2407.21783},
    archivePrefix={arXiv},
    primaryClass={cs.AI},
    url={https://arxiv.org/abs/2407.21783}
}

@inproceedings{mohri2024language,
    title={Language Models with Conformal Factuality Guarantees},
    author={Mohri, Christopher and Hashimoto, Tatsunori},
    booktitle={Proceedings of the 41st International Conference on Machine Learning},
    pages={36029--36047},
    year={2024},
    editor={Salakhutdinov, Ruslan and Kolter, Zico and Heller, Katherine and Weller, Adrian and Oliver, Nuria and Scarlett, Jonathan and Berkenkamp, Felix},
    volume={235},
    series={Proceedings of Machine Learning Research},
    month={21--27 Jul},
    publisher={PMLR},
    url={https://proceedings.mlr.press/v235/mohri24a.html}
}

@article{neyman1933problem,
    author={Neyman, Jerzy  and Pearson, Egon Sharpe},
    title={On the problem of the most efficient tests of statistical hypotheses},
    journal={Philosophical Transactions of the Royal Society of London. Series A},
    volume={231},
    number={694--706},
    pages={289--337},
    year={1933},
    doi={10.1098/rsta.1933.0009},
    url={https://royalsocietypublishing.org/doi/abs/10.1098/rsta.1933.0009},
    eprint={https://royalsocietypublishing.org/doi/pdf/10.1098/rsta.1933.0009}
}

@misc{openai2024gpt4,
    title={{GPT}-4 Technical Report},
    author={{OpenAI}},
    year={2024},
    eprint={2303.08774},
    archivePrefix={arXiv},
    primaryClass={cs.CL},
    url={https://arxiv.org/abs/2303.08774}
}

@misc{openai2024gpt4o,
    title={{GPT}-4o System Card},
    author={{OpenAI}},
    year={2024},
    eprint={2410.21276},
    archivePrefix={arXiv},
    primaryClass={cs.CL},
    url={https://arxiv.org/abs/2410.21276}
}

@misc{openai2024openaio1,
    title={{O}pen{AI} o1 System Card},
    author={{OpenAI}},
    year={2024},
    eprint={2412.16720},
    archivePrefix={arXiv},
    primaryClass={cs.AI},
    url={https://arxiv.org/abs/2412.16720}
}

@misc{qwen2025qwen2.5,
    title={{Q}wen2.5 Technical Report},
    author={{Qwen Team}},
    year={2025},
    eprint={2412.15115},
    archivePrefix={arXiv},
    primaryClass={cs.CL},
    url={https://arxiv.org/abs/2412.15115}
}

@article{ramdas2025hypothesis,
    url={http://dx.doi.org/10.1561/3600000002},
    year={2025},
    volume={1},
    journal={Foundations and Trends® in Statistics},
    title={Hypothesis Testing with E-values},
    doi={10.1561/3600000002},
    issn={2978-4212},
    number={1--2},
    pages={1--390},
    author={Aaditya Ramdas and Ruodu Wang}
}

@inproceedings{rubintoles2025conformal,
    title={Conformal Language Model Reasoning with Coherent Factuality},
    author={Maxon Rubin-Toles and Maya Gambhir and Keshav Ramji and Aaron Roth and Surbhi Goel},
    booktitle={The Thirteenth International Conference on Learning Representations},
    year={2025},
    url={https://openreview.net/forum?id=AJpUZd8Clb}
}

@article{shafer2021testing,
    author={Shafer, Glenn},
    title={Testing by Betting: A Strategy for Statistical and Scientific Communication},
    journal={Journal of the Royal Statistical Society Series A: Statistics in Society},
    volume={184},
    number={2},
    pages={407--431},
    year={2021},
    month={05},
    issn={0964-1998},
    doi={10.1111/rssa.12647},
    url={https://doi.org/10.1111/rssa.12647},
    eprint={https://academic.oup.com/jrsssa/article-pdf/184/2/407/49325712/jrsssa_184_2_407.pdf}
}

@article{shafer2008tutorial,
    author={Glenn Shafer and Vladimir Vovk},
    title={A Tutorial on Conformal Prediction},
    journal={Journal of Machine Learning Research},
    year={2008},
    volume={9},
    number={12},
    pages={371--421},
    url={http://jmlr.org/papers/v9/shafer08a.html}
}

@book{shafer2019gametheoretic,
    author={Glenn Shafer and Vladimir Vovk},
    publisher={John Wiley \& Sons, Ltd},
    isbn={9781118548035},
    title={Game‐Theoretic Foundations for Probability and Finance},
    doi={10.1002/9781118548035},
    url={https://onlinelibrary.wiley.com/doi/abs/10.1002/9781118548035},
    eprint={https://onlinelibrary.wiley.com/doi/pdf/10.1002/9781118548035},
    year={2019}
}

@misc{taori2023stanford,
    author={Rohan Taori and Ishaan Gulrajani and Tianyi Zhang and Yann Dubois and Xuechen Li and Carlos Guestrin and Percy Liang and Tatsunori B. Hashimoto},
    title={{S}tanford {A}lpaca: An Instruction-following {LL}a{MA} model},
    year={2023},
    publisher={GitHub},
    journal={GitHub repository},
    url={https://github.com/tatsu-lab/stanford_alpaca}
}

@misc{touvron2023llama2,
    title={{LL}a{MA} 2: Open Foundation and Fine-Tuned Chat Models},
    author={Hugo Touvron and Louis Martin and Kevin Stone and Peter Albert and Amjad Almahairi and Yasmine Babaei and Nikolay Bashlykov and Soumya Batra and Prajjwal Bhargava and Shruti Bhosale and Dan Bikel and Lukas Blecher and Cristian Canton Ferrer and Moya Chen and Guillem Cucurull and David Esiobu and Jude Fernandes and Jeremy Fu and Wenyin Fu and Brian Fuller and Cynthia Gao and Vedanuj Goswami and Naman Goyal and Anthony Hartshorn and Saghar Hosseini and Rui Hou and Hakan Inan and Marcin Kardas and Viktor Kerkez and Madian Khabsa and Isabel Kloumann and Artem Korenev and Punit Singh Koura and Marie-Anne Lachaux and Thibaut Lavril and Jenya Lee and Diana Liskovich and Yinghai Lu and Yuning Mao and Xavier Martinet and Todor Mihaylov and Pushkar Mishra and Igor Molybog and Yixin Nie and Andrew Poulton and Jeremy Reizenstein and Rashi Rungta and Kalyan Saladi and Alan Schelten and Ruan Silva and Eric Michael Smith and Ranjan Subramanian and Xiaoqing Ellen Tan and Binh Tang and Ross Taylor and Adina Williams and Jian Xiang Kuan and Puxin Xu and Zheng Yan and Iliyan Zarov and Yuchen Zhang and Angela Fan and Melanie Kambadur and Sharan Narang and Aurelien Rodriguez and Robert Stojnic and Sergey Edunov and Thomas Scialom},
    year={2023},
    eprint={2307.09288},
    archivePrefix={arXiv},
    primaryClass={cs.CL},
    url={https://arxiv.org/abs/2307.09288}
}

@article{tunstall2023creating,
    author={Tunstall, Lewis and Lambert, Nathan and Rajani, Nazneen and Beeching, Edward and Le Scao, Teven and von Werra, Leandro and Han, Sheon and Schmid, Philipp and Rush, Alexander},
    title={Creating a Coding Assistant with {S}tar{C}oder},
    journal={Hugging Face Blog},
    year={2023},
    url={https://huggingface.co/blog/starchat-alpha}
}

@article{vovk2025conformal,
    title={Conformal e-prediction},
    journal={Pattern Recognition},
    volume={166},
    pages={111674},
    year={2025},
    issn={0031-3203},
    doi={https://doi.org/10.1016/j.patcog.2025.111674},
    url={https://www.sciencedirect.com/science/article/pii/S0031320325003346},
    author={Vladimir Vovk}
}

@article{vovk2021evalues,
    author={Vladimir Vovk and Ruodu Wang},
    title={E-values: Calibration, combination and applications},
    volume={49},
    journal={The Annals of Statistics},
    number={3},
    publisher={Institute of Mathematical Statistics},
    pages={1736--1754},
    year={2021},
    doi={10.1214/20-AOS2020},
    url={https://doi.org/10.1214/20-AOS2020}
}

@book{vovk2022algorithmic,
    author={Vovk, Vladimir and Gammerman, Alexander and Shafer, Glenn},
    title={Algorithmic Learning in a Random World},
    year={2022},
    publisher={Springer International Publishing},
    address={Cham},
    isbn={978-3-031-06649-8},
    doi={10.1007/978-3-031-06649-8},
    url={https://doi.org/10.1007/978-3-031-06649-8}
}

@article{wald1939contributions,
    author={Abraham Wald},
    title={Contributions to the Theory of Statistical Estimation and Testing Hypotheses},
    volume={10},
    journal={The Annals of Mathematical Statistics},
    number={4},
    publisher={Institute of Mathematical Statistics},
    pages={299--326},
    year={1939},
    doi={10.1214/aoms/1177732144},
    url={https://doi.org/10.1214/aoms/1177732144}
}

@inproceedings{wang2024mathshepherd,
    title={{M}ath-{S}hepherd: Verify and Reinforce {LLM}s Step-by-step without Human Annotations},
    author={Wang, Peiyi and Li, Lei and Shao, Zhihong and Xu, Runxin and Dai, Damai and Li, Yifei and Chen, Deli and Wu, Yu and Sui, Zhifang},
    editor={Ku, Lun-Wei and Martins, Andre and Srikumar, Vivek},
    booktitle={Proceedings of the 62nd Annual Meeting of the Association for Computational Linguistics (Volume 1: Long Papers)},
    month=aug,
    year={2024},
    address={Bangkok, Thailand},
    publisher={Association for Computational Linguistics},
    url={https://aclanthology.org/2024.acl-long.510/},
    doi={10.18653/v1/2024.acl-long.510},
    pages={9426--9439}
}

@article{wang2022false,
    author={Wang, Ruodu and Ramdas, Aaditya},
    title={False Discovery Rate Control with E-values},
    journal={Journal of the Royal Statistical Society Series B: Statistical Methodology},
    volume={84},
    number={3},
    pages={822--852},
    year={2022},
    month={01},
    issn={1369-7412},
    doi={10.1111/rssb.12489},
    url={https://doi.org/10.1111/rssb.12489},
    eprint={https://academic.oup.com/jrsssb/article-pdf/84/3/822/49322390/jrsssb_84_3_822.pdf}
}

@article{wasserman2020universal,
    author={Larry Wasserman and Aaditya Ramdas and Sivaraman Balakrishnan},
    title={Universal inference},
    journal={Proceedings of the National Academy of Sciences},
    volume={117},
    number={29},
    pages={16880--16890},
    year={2020},
    doi={10.1073/pnas.1922664117},
    url={https://www.pnas.org/doi/abs/10.1073/pnas.1922664117},
    eprint={https://www.pnas.org/doi/pdf/10.1073/pnas.1922664117}
}

@inproceedings{wei2022chainofthought,
    author={Wei, Jason and Wang, Xuezhi and Schuurmans, Dale and Bosma, Maarten and ichter, brian and Xia, Fei and Chi, Ed and Le, Quoc V and Zhou, Denny},
    booktitle={Advances in Neural Information Processing Systems},
    editor={S. Koyejo and S. Mohamed and A. Agarwal and D. Belgrave and K. Cho and A. Oh},
    pages={24824--24837},
    publisher={Curran Associates, Inc.},
    title={Chain-of-Thought Prompting Elicits Reasoning in Large Language Models},
    url={https://proceedings.neurips.cc/paper_files/paper/2022/file/9d5609613524ecf4f15af0f7b31abca4-Paper-Conference.pdf},
    volume={35},
    year={2022}
}

@inproceedings{xu2024wizardlm,
    title={{W}izard{LM}: Empowering Large Pre-Trained Language Models to Follow Complex Instructions},
    author={Can Xu and Qingfeng Sun and Kai Zheng and Xiubo Geng and Pu Zhao and Jiazhan Feng and Chongyang Tao and Qingwei Lin and Daxin Jiang},
    booktitle={The Twelfth International Conference on Learning Representations},
    year={2024},
    url={https://openreview.net/forum?id=CfXh93NDgH}
}

@misc{yang2024qwen2,
    title={{Q}wen2 Technical Report},
    author={An Yang and Baosong Yang and Binyuan Hui and Bo Zheng and Bowen Yu and Chang Zhou and Chengpeng Li and Chengyuan Li and Dayiheng Liu and Fei Huang and Guanting Dong and Haoran Wei and Huan Lin and Jialong Tang and Jialin Wang and Jian Yang and Jianhong Tu and Jianwei Zhang and Jianxin Ma and Jianxin Yang and Jin Xu and Jingren Zhou and Jinze Bai and Jinzheng He and Junyang Lin and Kai Dang and Keming Lu and Keqin Chen and Kexin Yang and Mei Li and Mingfeng Xue and Na Ni and Pei Zhang and Peng Wang and Ru Peng and Rui Men and Ruize Gao and Runji Lin and Shijie Wang and Shuai Bai and Sinan Tan and Tianhang Zhu and Tianhao Li and Tianyu Liu and Wenbin Ge and Xiaodong Deng and Xiaohuan Zhou and Xingzhang Ren and Xinyu Zhang and Xipin Wei and Xuancheng Ren and Xuejing Liu and Yang Fan and Yang Yao and Yichang Zhang and Yu Wan and Yunfei Chu and Yuqiong Liu and Zeyu Cui and Zhenru Zhang and Zhifang Guo and Zhihao Fan},
    year={2024},
    eprint={2407.10671},
    archivePrefix={arXiv},
    primaryClass={cs.CL},
    url={https://arxiv.org/abs/2407.10671}
}

@misc{yang2024qwen2.5math,
    title={{Q}wen2.5-Math Technical Report: Toward Mathematical Expert Model via Self-Improvement},
    author={An Yang and Beichen Zhang and Binyuan Hui and Bofei Gao and Bowen Yu and Chengpeng Li and Dayiheng Liu and Jianhong Tu and Jingren Zhou and Junyang Lin and Keming Lu and Mingfeng Xue and Runji Lin and Tianyu Liu and Xingzhang Ren and Zhenru Zhang},
    year={2024},
    eprint={2409.12122},
    archivePrefix={arXiv},
    primaryClass={cs.CL},
    url={https://arxiv.org/abs/2409.12122}
}

@inproceedings{zheng2025processbench,
    title={{P}rocess{B}ench: Identifying Process Errors in Mathematical Reasoning},
    author={Zheng, Chujie and Zhang, Zhenru and Zhang, Beichen and Lin, Runji and Lu, Keming and Yu, Bowen and Liu, Dayiheng and Zhou, Jingren and Lin, Junyang},
    editor={Che, Wanxiang and Nabende, Joyce and Shutova, Ekaterina and Pilehvar, Mohammad Taher},
    booktitle={Proceedings of the 63rd Annual Meeting of the Association for Computational Linguistics (Volume 1: Long Papers)},
    month=jul,
    year={2025},
    address={Vienna, Austria},
    publisher={Association for Computational Linguistics},
    url={https://aclanthology.org/2025.acl-long.50/},
    doi={10.18653/v1/2025.acl-long.50},
    pages={1009--1024},
    isbn={979-8-89176-251-0}
}

\clearpage
\appendix
\onecolumn
%\thispagestyle{empty}
%\aistatstitle{\papertitle: \\ Supplementary Materials}
\section{PROOFS}
\label{appsec:theory}

We include the detailed derivation of \cref{theorem:size_distortion} below.
For convenience, we first re-state the theorem.

\theoremsizedistortion*

\begin{proof}

Note that a response is included in the filtered set $\mathds{S}_{\alpha} \rbr{x , g_{\pi} \rbr{x}}$ if and only if its score is $\leq \alpha$.
As a result, we can re-write the inclusion of an incorrect response as at least one incorrect response having a score $\leq \alpha$,
\begin{equation*}
    \exists \rbr{\mathbf{y} , \cdot} \in \mathds{S}_{\alpha} \rbr{x , g_{\pi} \rbr{x}} \text{ s.t. } o \rbr{x , \mathbf{y}} = 0 \iff \min_{\rbr{\mathbf{y} , c} \in \mathds{O} \rbr{x , g_{\pi} \rbr{x}} : c = 0} s \rbr{x , \mathbf{y}} \leq \alpha .
\end{equation*}
Therefore, the size distortion for any post-hoc $\alpha$ strategy is upper bound by the following (simplified) worst-case,
\begin{equation}
    \label{eq:max_size_distortion}
    \begin{gathered}
        \ev{\frac{\ind{\exists \rbr{\mathbf{Y} , \cdot} \in \mathds{S}_{\alpha \rbr{X , \mathds{S} \rbr{X , g_{\pi} \rbr{X}}}} \rbr{X , g_{\pi} \rbr{X}} \text{ s.t. } o \rbr{X , \mathbf{Y}} = 0}}{\alpha \rbr{X , \mathds{S} \rbr{X , g_{\pi} \rbr{X}}}}} \\
        \leq \ev{\max_{\alpha \in \mathds{R}_{\geq 0}} \frac{\ind{\exists \rbr{\mathbf{Y} , \cdot} \in \mathds{S}_{\alpha} \rbr{X , g_{\pi} \rbr{X}} \text{ s.t. } o \rbr{X , \mathbf{Y}} = 0}}{\alpha}} \\
        = \ev{\max_{\alpha \in \mathds{R}_{\geq 0}} \frac{\ind{\min_{\rbr{\mathbf{Y} , C} \in \mathds{O} \rbr{X , g_{\pi} \rbr{X}} : C = 0} s \rbr{X , \mathbf{Y}} \leq \alpha}}{\alpha}} \\
        \overset{(i)}{=} \ev{\rbr{\min_{\rbr{\mathbf{Y} , C} \in \mathds{O} \rbr{X , g_{\pi} \rbr{X}} : C = 0} s \rbr{X , \mathbf{Y}}}^{- 1}} = \ev{\max_{\rbr{\mathbf{Y} , C} \in \mathds{O} \rbr{X , g_{\pi} \rbr{X}} : C = 0} \rbr{s \rbr{X , \mathbf{Y}}}^{- 1}} ,
    \end{gathered}
\end{equation}
where the equality $(i)$ is achieved by setting $\alpha$ to the smallest value for which the indicator function evaluates to 1, otherwise the whole term is 0.
We are interested in upper bounding the above expectation by 1 to satisfy \cref{eq:size_distortion}.

Now, we use the definition of our proposed e-scores from \cref{eq:e_score}.
Note that our e-scores depend on the calibration data; we will make this dependence explicit in the following.
Then, the worst-case size distortion simplifies to,
\begin{equation*}
    \begin{gathered}
        \ev{\max_{\rbr{\mathbf{Y}^{n + 1} , C^{n + 1}} \in \mathds{O} \rbr{X^{n + 1} , g_{\pi} \rbr{X^{n + 1}}} : C^{n + 1} = 0} \rbr{s_{\text{e-score}} \rbr{X^{n + 1} , \mathbf{Y}^{n + 1} ; X^{1} , \ldots , X^{n}}}^{- 1}} \\
        = \ev{\max_{\rbr{\mathbf{Y}^{n + 1} , C^{n + 1}} \in \mathds{O} \rbr{X^{n + 1} , g_{\pi} \rbr{X^{n + 1}}} : C^{n + 1} = 0} \frac{\rbr{n + 1} \cdot f \rbr{X^{n + 1} , \mathbf{Y}^{n + 1}}}{f \rbr{X^{n + 1} , \mathbf{Y}^{n + 1}} + \sum_{i = 1}^{n} f^{*} \rbr{X^{i} , \mathds{O} \rbr{X^{i} , g_{\pi} \rbr{X^{i}}}}}} .
    \end{gathered}
\end{equation*}
Note that for $a, b \in \mathds{R}_{\geq 0}$, the ratio $\nicefrac{a}{\rbr{a + b}}$ is a monotonically non-decreasing transformation of $a$ because the derivative with respect to $a$ (i.e., $\nicefrac{b}{\rbr{a + b}^{2}}$) is non-negative.
Consequently, the above maximum is achieved at $f^{*} \rbr{x , \mathds{O} \rbr{x , g_{\pi} \rbr{x}}} = \max_{\rbr{\mathbf{y} , c} \in \mathds{O} \rbr{x , g_{\pi} \rbr{x}} : c = 0} f \rbr{x , \mathbf{y}}$.
Therefore, the worst-case size distortion simplifies to,
\begin{equation*}
    \begin{gathered}
        \ev{\max_{\rbr{\mathbf{Y}^{n + 1} , C^{n + 1}} \in \mathds{O} \rbr{X^{n + 1} , g_{\pi} \rbr{X^{n + 1}}} : C^{n + 1} = 0} \frac{\rbr{n + 1} \cdot f \rbr{X^{n + 1} , \mathbf{Y}^{n + 1}}}{f \rbr{X^{n + 1} , \mathbf{Y}^{n + 1}} + \sum_{i = 1}^{n} f^{*} \rbr{X^{i} , \mathds{O} \rbr{X^{i} , g_{\pi} \rbr{X^{i}}}}}} \\
        = \ev{\frac{\rbr{n + 1} \cdot f^{*} \rbr{X^{n + 1} , \mathds{O} \rbr{X^{n + 1} , g_{\pi} \rbr{X^{n + 1}}}}}{f^{*} \rbr{X^{n + 1} , \mathds{O} \rbr{X^{n + 1} , g_{\pi} \rbr{X^{n + 1}}}} + \sum_{i = 1}^{n} f^{*} \rbr{X^{i} , \mathds{O} \rbr{X^{i} , g_{\pi} \rbr{X^{i}}}}}} \\
        = \ev{\frac{\rbr{n + 1} \cdot f^{*} \rbr{X^{n + 1} , \mathds{O} \rbr{X^{n + 1} , g_{\pi} \rbr{X^{n + 1}}}}}{\sum_{i = 1}^{n + 1} f^{*} \rbr{X^{i} , \mathds{O} \rbr{X^{i} , g_{\pi} \rbr{X^{i}}}}}} .
    \end{gathered}
\end{equation*}

Lastly, we assume that the test and the calibration prompts are exchangeable, i.e., for any permutation $\sigma$ over the indices $\cbr{1 , \ldots , n + 1}$, the ordering of the permuted prompts is equal in distribution to the un-permuted prompts,
\begin{equation*}
    \rbr{X^{\sigma \rbr{1}} , \ldots , X^{\sigma \rbr{n + 1}}} \overset{d}{=} \rbr{X^{1} , \ldots , X^{n + 1}} .
\end{equation*}
We can follow arguments similar to those made by \citet{gammerman1998learning,balinsky2024enhancing,vovk2025conformal} to show that the above expectation is $\leq 1$ under exchangeability.
Specifically, we define random variables,
\begin{equation*}
    R^{i} = \frac{\rbr{n + 1} \cdot f^{*} \rbr{X^{i} , \mathds{O} \rbr{X^{i} , g_{\pi} \rbr{X^{i}}}}}{\sum_{j = 1}^{n + 1} f^{*} \rbr{X^{j} , \mathds{O} \rbr{X^{j} , g_{\pi} \rbr{X^{j}}}}} ,
\end{equation*}
for $i = 1 , \ldots , n + 1$.
Under exchangeability of $X^{1} , \ldots X^{n + 1}$, the distributions of $R^{1} , \ldots R^{n + 1}$ are identical.
Then,
\begin{equation*}
    \begin{gathered}
        \ev{\frac{\rbr{n + 1} \cdot f^{*} \rbr{X^{n + 1} , \mathds{O} \rbr{X^{n + 1} , g_{\pi} \rbr{X^{n + 1}}}}}{\sum_{i = 1}^{n + 1} f^{*} \rbr{X^{i} , \mathds{O} \rbr{X^{i} , g_{\pi} \rbr{X^{i}}}}}} = \ev{R^{n + 1}} = \frac{\sum_{i = 1}^{n + 1} \ev{R^{i}}}{n + 1} = \frac{\ev{\sum_{i = 1}^{n + 1} R^{i}}}{n + 1} \\
        = \frac{\ev{\sum_{i = 1}^{n + 1} \frac{\rbr{n + 1} \cdot f^{*} \rbr{X^{i} , \mathds{O} \rbr{X^{i} , g_{\pi} \rbr{X^{i}}}}}{\sum_{j = 1}^{n + 1} f^{*} \rbr{X^{j} , \mathds{O} \rbr{X^{j} , g_{\pi} \rbr{X^{j}}}}}}}{n + 1} = \ev{\frac{\sum_{i = 1}^{n + 1} f^{*} \rbr{X^{i} , \mathds{O} \rbr{X^{i} , g_{\pi} \rbr{X^{i}}}}}{\sum_{j = 1}^{n + 1} f^{*} \rbr{X^{j} , \mathds{O} \rbr{X^{j} , g_{\pi} \rbr{X^{j}}}}}} \overset{(ii)}{\leq} \ev{1} = 1 ,
    \end{gathered}
\end{equation*}
where the inequality $(ii)$ accounts for the sum being 0, making $\nicefrac{0}{0} = 0$ (by convention).
Hence, our e-scores in \cref{eq:e_score} upper bound the size distortion (marginal over the test and the calibration prompts) by 1, as in \cref{eq:size_distortion}.

Furthermore, we can use the definition of our proposed combined e-scores from \cref{eq:combined_e_score}.
Note that instead of combining three e-scores, we can combine any $k \geq 1$.
The worst-case size distortion from \cref{eq:max_size_distortion} simplifies to,
\begin{equation*}
    \begin{gathered}
        \ev{\max_{\rbr{\mathbf{Y}^{n + 1} , C^{n + 1}} \in \mathds{O} \rbr{X^{n + 1} , g_{\pi} \rbr{X^{n + 1}}} : C^{n + 1} = 0} \rbr{s_{\text{e-score (combined)}} \rbr{X^{n + 1} , \mathbf{Y}^{n + 1} ; X^{1} , \ldots , X^{n}}}^{- 1}} \\
        = \ev{\max_{\rbr{\mathbf{Y}^{n + 1} , C^{n + 1}} \in \mathds{O} \rbr{X^{n + 1} , g_{\pi} \rbr{X^{n + 1}}} : C^{n + 1} = 0} \frac{\sum_{i = 1}^{k} \rbr{s_{\text{e-score ($i$)}} \rbr{X^{n + 1} , \mathbf{Y}^{n + 1} ; X^{1} , \ldots , X^{n}}}^{- 1}}{k}} \\
        \leq \frac{\sum_{i = 1}^{k} \ev{\max_{\rbr{\mathbf{Y}^{n + 1} , C^{n + 1}} \in \mathds{O} \rbr{X^{n + 1} , g_{\pi} \rbr{X^{n + 1}}} : C^{n + 1} = 0} \rbr{s_{\text{e-score ($i$)}} \rbr{X^{n + 1} , \mathbf{Y}^{n + 1} ; X^{1} , \ldots , X^{n}}}^{- 1}}}{k} .
    \end{gathered}
\end{equation*}
We have shown that the worst-case size distortion for individual e-scores (in the numerator) is $\leq 1$. Then,
\begin{equation*}
    \frac{\sum_{i = 1}^{k} \ev{\max_{\rbr{\mathbf{Y}^{n + 1} , C^{n + 1}} \in \mathds{O} \rbr{X^{n + 1} , g_{\pi} \rbr{X^{n + 1}}} : C^{n + 1} = 0} \rbr{s_{\text{e-score ($i$)}} \rbr{X^{n + 1} , \mathbf{Y}^{n + 1} ; X^{1} , \ldots , X^{n}}}^{- 1}}}{k} \leq \frac{\sum_{i = 1}^{k} 1}{k} = 1 .
\end{equation*}
Hence, our combined e-scores in \cref{eq:combined_e_score} upper bound the size distortion (marginally) by 1, as in \cref{eq:size_distortion}.

\end{proof}

\section{EXPERIMENTAL RESULTS FOR WORST-CASE ANALYSIS}
\label{appsec:experiments}

We include additional experimental results here, expanding on \cref{sec:experiments}.
In particular, we perform a worst-case analysis for the different use-cases, where the post-hoc $\alpha$ strategy maximizes size distortion, which we quantified in \cref{eq:max_size_distortion}.
We begin by stating the common baselines considered, in addition to the p-scores from \cref{sec:experiments}.

\paragraph{Baselines}
In addition to the p-scores defined in \cref{eq:p_score}, we also compare with their randomized version,
\begin{equation*}
    \begin{gathered}
        s_{\text{p-score (randomized)}} \rbr{x^{n + 1} , \mathbf{y}^{n + 1}} = \\
        \frac{u \cdot \rbr{1 + \sum_{i = 1}^{n} \ind{\splitfrac{f \rbr{x^{n + 1} , \mathbf{y}^{n + 1}}}{= f^{*} \rbr{x^{i} , \mathds{O} \rbr{x^{i} , g_{\pi} \rbr{x^{i}}}}}}} + \sum_{i = 1}^{n} \ind{\splitfrac{f \rbr{x^{n + 1} , \mathbf{y}^{n + 1}}}{< f^{*} \rbr{x^{i} , \mathds{O} \rbr{x^{i} , g_{\pi} \rbr{x^{i}}}}}}}{n + 1} ,
    \end{gathered}
\end{equation*}
where $u \sim \mathcal{U} \rbr{0 , 1}$ is a uniform random sample in the range $\sbr{0 , 1}$.
We can recover the p-scores defined in \cref{eq:p_score} as a special case of this definition by deterministically setting $u = 1$.
While the non-randomized p-scores correspond to p-values, these randomized p-scores correspond to exact p-values \citep{shafer2008tutorial,vovk2022algorithmic}.\footnote{Consider a non-negative random variable $R \in \mathds{R}_{\geq 0}$.
It is an \emph{exact p-variable} if $\pr{R \leq \alpha} = \alpha$, for all $\alpha \in \sbr{0 , 1}$.}

We also compare with the transformed oracle estimators in \cref{eq:f_options} directly, without any conversion to e- or p-scores using the calibration data.
Since we want our scores to be low for correct and high for incorrect responses (as measures of incorrectness), we define the \emph{naive} scores to be the reciprocal of the transformed oracle estimators,
\begin{equation*}
    s_{\text{naive}} \rbr{x^{n + 1} , \mathbf{y}^{n + 1}} \! = \! \rbr{f_{\hat{o}} \rbr{x^{n + 1} , \mathbf{y}^{n + 1}}}^{- 1} \! = \! \begin{cases}
        \rbr{\hat{o} \rbr{x^{n + 1} , \mathbf{y}^{n + 1}}}^{- 1} \in \sbr{1 , \infty} & \text{(for naive 1)} \\
        1 - \hat{o} \rbr{x^{n + 1} , \mathbf{y}^{n + 1}} \in \sbr{0 , 1} & \text{(for naive 2)} \\
        \rbr{\hat{o} \rbr{x^{n + 1} , \mathbf{y}^{n + 1}}}^{- 1} \cdot \rbr{1 - \hat{o} \rbr{x^{n + 1} , \mathbf{y}^{n + 1}}} \in \sbr{0 , \infty} & \text{(for naive 3)}
    \end{cases} .
\end{equation*}
These naive scores generally do not come with any statistical guarantees by themselves.
However, because the reciprocal of naive (1) is always $\leq 1$, it happens to correspond to an \emph{uninformative} e-value that is always $\leq 1$ (the expectation is $\leq 1$ by design).
Therefore, even though naive (1) achieves the size distortion bound in \cref{eq:size_distortion}, it regularly excludes responses (correct and incorrect) and is extremely conservative compared to our e-scores.

\subsection{Worst-Case Size Distortion Analysis}
\label{appsubsec:experiments-worst}

\begin{table}[!t]
    \caption{
        \textbf{Scores for the worst-case size distortion analysis.}
        We use the mathematical factuality (cf. \cref{subsec:experiments-processbench}) and property constraints satisfaction (cf. \cref{subsec:experiments-ultrafeedback}) settings to compare our proposed e-scores against p-scores and naive scores.
        We consider the worst-case that maximizes size distortion (cf. \cref{eq:max_size_distortion}).
        We report the mean and the inter-quartile range (which depicts the 25-th and 75-th quantiles) of the size distortion.
    }
    \label{tab:worst_case}
    \begin{center}
        \begin{small}
            \begin{tabular}{L{3cm}C{3cm}C{3cm}C{3cm}C{3cm}}
                \toprule
                \textbf{Score} & \multicolumn{4}{c}{\textbf{Worst-case size distortion}} \\
                \cmidrule(lr){2-5}
                & \multicolumn{2}{c}{\textbf{Mathematical factuality}} & \multicolumn{2}{c}{\textbf{Property constraints satisfaction}} \\
                \cmidrule(lr){2-3} \cmidrule(lr){4-5}
                & \textbf{QwenPRM} & \textbf{MathShepherd-PRM} & \textbf{Helpfulness and instruction-following} & \textbf{Honesty and truthfulness} \\
                \midrule
                naive (1) & 0.24\textsubscript{(0.00-0.46)} & 0.20\textsubscript{(0.00-0.37)} & 0.07\textsubscript{(0.00-0.01)} & 0.09\textsubscript{(0.00-0.03)} \\
                naive (2) & 1.89\textsubscript{(0.00-1.86)} & 1.30\textsubscript{(0.00-1.59)} & 2.25\textsubscript{(0.00-1.01)} & 5.42\textsubscript{(0.00-1.03)} \\
                naive (3) & 1.39\textsubscript{(0.00-0.86)} & 0.80\textsubscript{(0.00-0.59)} & 1.80\textsubscript{(0.00-0.01)} & 4.82\textsubscript{(0.00-0.03)} \\
                \midrule
                p-score & 7.21\textsubscript{(0.00-4.00)} & 7.46\textsubscript{(0.00-4.03)} & 9.60\textsubscript{(0.00-4.00)} & 7.55\textsubscript{(0.00-3.98)} \\
                p-score (randomized) & 15.80\textsubscript{(0.00-4.00)} & 13.96\textsubscript{(0.00-4.03)} & 15.91\textsubscript{(0.00-4.00)} & 14.92\textsubscript{(0.00-3.99)} \\
                \midrule
                e-score (1) & 1.00\textsubscript{(0.00-1.95)} & 1.01\textsubscript{(0.00-1.90)} & 1.00\textsubscript{(0.00-0.19)} & 1.00\textsubscript{(0.00-0.34)} \\
                e-score (2) & 0.79\textsubscript{(0.00-0.79)} & 0.73\textsubscript{(0.00-0.89)} & 0.80\textsubscript{(0.00-0.38)} & 0.97\textsubscript{(0.00-0.23)} \\
                e-score (3) & 1.01\textsubscript{(0.00-0.63)} & 1.01\textsubscript{(0.00-0.75)} & 1.00\textsubscript{(0.00-0.01)} & 1.05\textsubscript{(0.00-0.01)} \\
                e-score (combined) & 0.94\textsubscript{(0.00-1.12)} & 0.91\textsubscript{(0.00-1.18)} & 0.94\textsubscript{(0.00-0.19)} & 1.01\textsubscript{(0.00-0.19)} \\
                \bottomrule
            \end{tabular}
        \end{small}
    \end{center}
\end{table}

We analyze the worst-case size distortion in \cref{eq:max_size_distortion}.
\cref{tab:worst_case} illustrates the results for both our experimental use-cases: mathematical factuality (cf. \cref{subsec:experiments-processbench}) and property constraints satisfaction (cf. \cref{subsec:experiments-ultrafeedback}).
Our proposed e-scores (and naive (1)) reliably upper bound the worst-case size distortion to 1 and satisfy \cref{eq:size_distortion}, corroborating our theory in \cref{theorem:size_distortion}.
Conversely, p-scores and other naive scores are unable to achieve this.

\section{IMPLICIT P-SCORES IN RELATED WORK}
\label{appsec:related}

Here we highlight the implicit role of p-scores (cf. \cref{eq:p_score}), and hence p-values, in the works most closely related to ours \citep{mohri2024language,cherian2024large,rubintoles2025conformal}, making it explicit.
To begin with, these works compute the calibration values $f^{*} \rbr{x^{i} , \mathds{O} \rbr{x^{i} , g_{\pi} \rbr{x^{i}}}}$ for $i = 1 , \ldots , n$.
Given a fixed user-defined $\alpha \in \sbr{\nicefrac{1}{\rbr{n + 1}} , 1}$, they compute a threshold $\tau_{\alpha}$ set to the $\lceil \rbr{1 - \alpha} \cdot \rbr{n + 1} \rceil$-th smallest of the calibration values above.
Then, a test response $\mathbf{y}^{n + 1}$ is included in the returned set if $f \rbr{x^{n + 1} , \mathbf{y}^{n + 1}}$ is larger than this threshold,
\begin{equation*}
    \begin{gathered}
        f \rbr{x^{n + 1} , \mathbf{y}^{n + 1}} > \tau_{\alpha} \\
        \iff \sum_{i = 1}^{n} \ind{f \rbr{x^{n + 1} , \mathbf{y}^{n + 1}} > f^{*} \rbr{x^{i} , \mathds{O} \rbr{x^{i} , g_{\pi} \rbr{x^{i}}}}} \geq \rbr{1 - \alpha} \cdot \rbr{n + 1} \\
        \iff \sum_{i = 1}^{n} \ind{f \rbr{x^{n + 1} , \mathbf{y}^{n + 1}} \leq f^{*} \rbr{x^{i} , \mathds{O} \rbr{x^{i} , g_{\pi} \rbr{x^{i}}}}} \leq \alpha \cdot \rbr{n + 1} - 1 \\
        \iff \frac{1 + \sum_{i = 1}^{n} \ind{f \rbr{x^{n + 1} , \mathbf{y}^{n + 1}} \leq f^{*} \rbr{x^{i} , \mathds{O} \rbr{x^{i} , g_{\pi} \rbr{x^{i}}}}}}{n + 1} \leq \alpha \\
        \iff s_{\text{p-score}} \rbr{x^{n + 1} , \mathbf{y}^{n + 1}} \leq \alpha .
    \end{gathered}
\end{equation*}
In our setup, this is equivalent to returning the filtered set $\mathds{S}_{\alpha} \rbr{x^{n + 1} , g_{\pi} \rbr{x^{n + 1}}}$ using p-scores.
We again highlight that these approaches satisfy \cref{eq:size}, but not its post-hoc generalization in \cref{eq:size_distortion}; for that, we propose e-scores.

\end{document}